\documentclass{article}
\usepackage{iclr2027_conference,times}
\usepackage{flafter}

\usepackage[utf8]{inputenc}
\usepackage[T1]{fontenc}
\usepackage[hidelinks]{hyperref}   %
\usepackage{url}
\usepackage{booktabs}
\usepackage{amsfonts}
\usepackage{nicefrac}
\usepackage{microtype}
\usepackage{graphicx}
\usepackage[table]{xcolor}
\usepackage[most]{tcolorbox}
\newtcolorbox{promptbox}[1]{colback=black!4, colframe=black!45, boxrule=0.4pt,
  arc=1.2pt, left=5pt, right=5pt, top=3pt, bottom=3pt, boxsep=1pt,
  fonttitle=\small\bfseries, coltitle=black, colbacktitle=black!12,
  title={#1}}
\usepackage{multirow}
\usepackage{amsmath}
\usepackage{xcolor}
\usepackage{subcaption}

\title{Retrieved but Not Delivered:\\Multimodal Memory Delivery\\for Long-Term Agents}

\author{Yuhang Jiang\thanks{Corresponding author. Work done during a research internship at Huawei Pisa Research Center.}\kern4pt,
Qingwei Liao, Kaize Yin, Xingling Liu, Luca Cuomo, Silvio Bacci \\
Huawei Technologies Ltd. \\
{\small \texttt{jyhtjtj@gmail.com}, \texttt{\{liaoqingwei,yinkaize,l.cuomo,s.bacci\}@huawei.com}} \\
{\small \texttt{liuxingling5@h-partners.com}}}

\iclrfinalcopy

\begin{document}
\maketitle
\lhead{Preprint}

\begin{abstract}
Work on memory for multimodal agents optimizes what is written, updated and retrieved. Between retrieval and the answer,
however, is a stage that multimodal memory evaluations do not isolate: what of the retrieved
memory reaches the model, and in what form. We call it delivery, and a controlled
decomposition on MemLens locates the remaining room there. With the retrieved evidence
set exactly fixed, delivering the original pixels instead of withholding them raises accuracy by
13.87 points on an 8B backbone, whereas making retrieval perfect on those same messages
improves it by 2.31. Delivery is the larger term on all three MemLens backbones and grows
with backbone strength; retrieval grows too, without closing the gap. We propose \textbf{DeliverMem}, an instantiation of delivery as three decisions:
keep the original modality, give each item a readable identity, and state when it was seen,
with a retrieval-side adapter for the one property delivery cannot supply.
Each is measured against a delivery-matched control that alters only its own variable.
DeliverMem leads the strongest published memory agent on MemLens at all four context lengths,
and beats DMV-Bench's own strongest method at every setting on both backbones. On MemLens it
does this on a tenth to a seventieth of the input. Each decision helps only where the question
lacks what it supplies, and is null elsewhere.
A single fixed configuration nonetheless leads both benchmarks, without training any component
or modifying the stored records. Project page: \url{https://avalon-s.github.io/DeliverMem/}
\end{abstract}

\begin{figure}[!ht]
\centering
\includegraphics[width=\textwidth]{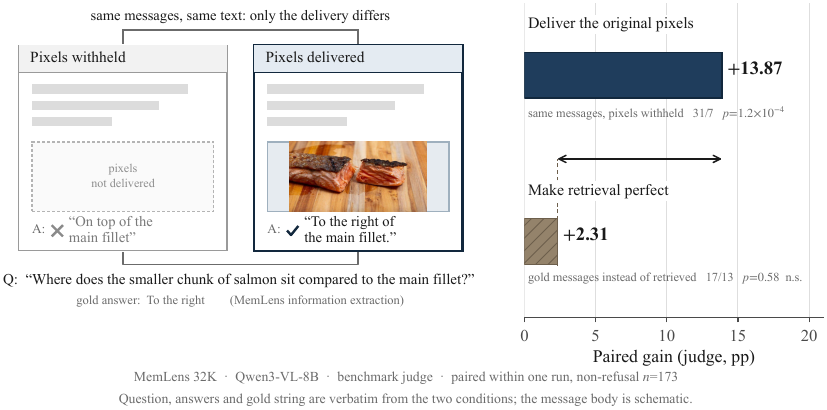}
\caption{Left: one MemLens question from our runs, answered with the pixels withheld and
with them delivered. Right: the two quantities this paper separates, with the arrow marking
the difference between them.}
\label{fig:teaser}
\end{figure}

\section{Introduction}\label{sec:intro}

Users expect an AI assistant to remember the photographs they send as well as what they
write. One user, testing a citrus-cured salmon recipe, sends a photograph and asks for a
calm, sensory description to adapt into a cookbook intro. Later comes a question: where does
the smaller chunk of salmon sit compared to the main fillet? Nothing the user wrote mentions
it, but the photograph shows it. The exchange is from MemLens \citep{memlens}, a benchmark of
questions about long multimodal conversations. Our retriever finds the message and places it
in the prompt, and with its pixels Qwen3-VL-8B replies ``to the right of the
main fillet''. Without its pixels, as text-only memory would deliver it, the model replies
``on top of the main fillet'', a confident wrong answer (Figure~\ref{fig:teaser}). Retrieval has done its job, and the answer still turns on the form
in which the evidence arrives.

An agent that operates over weeks accumulates far more history than fits in a context
window, and the standard response is a memory system built on retrieval-augmented
generation \citep{rag}: experience is written to a store $\mathcal{S}$, a retriever $R$
returns a small set of records per query, and the answering model $M$ conditions on them.
Between the last two is a step $D$ that turns retrieved records into
the tokens and pixels placed in the prompt, making the answer
$a = M(q, D(R(q,\mathcal{S})))$.
Research has concentrated on $\mathcal{S}$ and $R$ \citep{memgpt,genagents,mem0,amem}. In
text-only memory, $D$ is easy to fold into retrieval: the retrieved strings are the delivered
strings, and the memory literature grew up text-first. A systems characterisation of agent
memory names the stage but registers it as a prefill cost rather than a term in accuracy
\citep{agentmemsys}. The strongest published system on the visual-memory benchmark DMV-Bench
\citep{dmvbench} stores a visual and a verbal code, following dual-coding theory
\citep{paivio1971}, and nothing in that theory says which of the two to deliver.

This paper is about $D$, the step we call delivery. Retrieval decides which memories
are accessible. Delivery decides which information in them is usable by the
answering model. In multimodal memory a visual memory can arrive as pixels, a caption
or a crop, and the model treats each differently. A controlled decomposition on
MemLens measures delivery against retrieval: we feed the model the gold
messages for each question, leaving retrieval at ceiling, and vary only whether their images
are delivered. Delivering them improves accuracy by $13.87$ points. Replacing our retrieved
set with the gold set, on the same questions and judge, measures the effect of perfect
retrieval: $+2.31$ points, indistinguishable from zero (Figure~\ref{fig:teaser}). Delivery is
the larger term on all three Qwen3-VL backbones we test, significantly so at $8$B, and grows
as the backbone does (Section~\ref{sec:decomp}).

Beyond modality, delivery decides which item is which (identity), when it was seen
(temporal position), which image belongs to which message (binding), in what
order, and how many. We measure the first three, fix order in every control, and vary how
many in Section~\ref{sec:decomp}. One objection is that delivery is formatting, and every
retrieval system formats its prompt. Controls that fix the delivered set and its order show
otherwise: accuracy moves with the content of the pixels, which image follows which
message, and whether an identifier sits in the pixels or in the text beside them
(Section~\ref{sec:controls}). Their results follow one rule: a
delivery decision helps where (i)~the information survives storage and reaches the model,
(ii)~the question needs the property the decision supplies, and (iii)~the delivered context
does not already carry it; it is null otherwise. Condition~(i) is what makes this a stage and
not a system: the pipeline is serial, and no way of presenting an item recovers one that
was never retrieved. Every null we measure is one of those three conditions failing. What
helps therefore depends on what the question asks, and one fixed configuration that applies
every decision to every question leads on both benchmarks.

\paragraph{Contributions.}
\begin{enumerate}\itemsep2pt
\item We separate delivery from retrieval with a controlled decomposition, and
measure each delivery decision against a
delivery-matched control that changes only that decision's variable. Relative to
current practice on each side, delivery has more headroom on MemLens than retrieval
(Section~\ref{sec:decomp}).
\item We propose DeliverMem, an interface that keeps the original modality, gives
each item a readable identity and states when it was seen, with a retrieval-side adapter
for the property delivery cannot supply. Nothing is trained and no stored record is modified (Section~\ref{sec:method}).
\item DeliverMem leads every published memory agent on MemLens at all four
context lengths,
from an $8$B open model delivering ten messages, and beats DMV-Bench's strongest
method at every setting on two backbones. Its $32$K score matches
a frontier proprietary large vision-language model (LVLM) reading the whole conversation, on a tenth of the input
(Section~\ref{sec:results}).
\end{enumerate}

\section{Related Work}\label{sec:related}

Surveys organise agent memory by the stage a mechanism acts on \citep{memsurvey}, and work
has concentrated on three: what to persist and in what structure
\citep{memgpt,genagents,mem0,amem}, how stored facts are revised or forgotten
\citep{supersede}, and how the relevant subset is found, including
graph-structured stores and late-interaction multi-vector matching \citep{colpali,argus}.

Systems that caption on ingestion discard the pixels \citep{mirix,memverse}, and the penalty grows with how fine-grained the required evidence is \citep{memeye}. That loss is taken at write time, and Section~\ref{sec:decomp} measures it at delivery, with the pixels still in the store. Pixels alone are not enough: LVLMs largely fail to notice when an
image is swapped for a similar but semantically different one \citep{illusion}, and
whole-screen visual memory in GUI agents reduces state-level failures while worsening
action-level ones \citep{naivevis}. Delivered items must also be addressable.

Our identity tag relocates Set-of-Mark prompting \citep{som} from regions of one image to items of a retrieved set, and our retrieval adapter scores whole and regional SigLIP~2 \citep{siglip2} encodings by MaxSim \citep{colbert}. Recent papers act on adjacent decisions. OCR-Memory
\citep{ocrmemory} gives rendered trajectories unique visual identifiers, but as part of a
retrieval representation rather than at delivery, and without a control separating
uniqueness from salience. PMMC \citep{pmmc} moves reasoning to consolidation time,
changing what is stored, whereas here the stored and retrieved sets stay fixed.
MemOCR \citep{memocr} renders a structured text memory into an image and trains a
budget-aware policy over it, so its pixels are a compressed rendering of text chosen to
save tokens; the decision it optimises is how much to compress, not whether the original
record survives to delivery.
SMMBench \citep{smmbench} is a benchmark built around evidence scattered across sources.
The three systems change what is stored or how it is represented for retrieval, and the benchmark evaluates those stages; none of the four holds the retrieved set fixed and varies only what reaches the model.

MemGPT \citep{memgpt} and MemOS \citep{memos} page content between a working context and
external tiers; Mem0 \citep{mem0} and A-Mem \citep{amem} maintain an evolving store of
extracted facts; MemAgent \citep{memagent} trains an overwrite policy over segments;
Memory-T1 \citep{memoryt1} learns which sessions to select; M3-Agent \citep{m3agent} is
multimodal and builds an entity-centric graph over video and audio. All of them decide at write and retrieve time and leave the form a record takes between retrieval and the prompt to the implementation.
Whether a system keeps its pixels depends on the unit of memory it stores, not
on whether the agent is embodied (Appendix~\ref{app:unit}).
The post-retrieval stage is beginning to be formalised in text-only agents
\citep{postretrieval,diagnosing,memarbiter,actmem}, where no modality decision arises. The nearest multimodal systems act on
retrieval rather than delivery \citep{mmexam,vmem}. Appendix~\ref{app:related}
places this paper against both, and against the presentation and encoder confounds
behind our controls \citep{presentation,memdelta}.

\section{Benchmarks}\label{sec:benchmarks}

We evaluate on two public benchmarks for multimodal agent memory, neither built for this
paper.

\paragraph{DMV-Bench: image-level visual recall.}
DMV-Bench \citep{dmvbench} is an interactive benchmark for an agent's visual
memory: an agent browses a simulated storefront across $J$ sessions, one product
photograph carries a small object that appears nowhere in the page text, and sessions later the agent must return to the product that carried it. The answer is a specific image,
beyond the reach of a system that stores only the page text. Scoring is exact, with no judge model. We use both
backbones the benchmark reports, Qwen2.5-VL-7B \citep{qwen25vl} and
Gemini~2.5~Flash \citep{gemini25}, and run its dual-bank method, DualMem, from the original repository at the authors' best reported setting (Appendix~\ref{app:setup}).

\paragraph{MemLens: long-context multimodal conversational memory.}
MemLens \citep{memlens} pairs a long multimodal conversation with questions whose answers
are supported by a small number of messages scattered through it, across four released
context lengths. Its own image ablation reports that removing the evidence images drops two frontier models below $2\%$ accuracy on the
$80.4\%$ of questions whose evidence includes an image. We use the canonical
$195$-question subset, an $8$B backbone and the benchmark's own $235$B judge,
both from the Qwen3-VL family \citep{qwen3vl}. Its five question types are information extraction, knowledge update, temporal reasoning, multi-session reasoning and answer refusal, the last a test of abstention (Appendix~\ref{app:setup}). The benchmark separates two method families,
long-context LVLMs that read everything and memory-augmented agents that read a selection.
Because the families differ by more than an order of magnitude in inference cost, we compare
against the memory-augmented agents and report the long-context models alongside them.
Appendix~\ref{app:pool} gives the 32K pool in full and names the best agent at each length, and
Appendix~\ref{app:refsys} states why we take the best overall system as the
reference point on each.
The two benchmarks are complementary: DMV-Bench asks for a single decisive image, MemLens for a fact
spread over several messages, and a method that worked only on DMV-Bench would be a
image-retrieval method.

\section{Separating Delivery from Retrieval}\label{sec:decomp}

All contrasts below are paired per question (Appendix~\ref{app:batches}): the answering model, the prompt template and the judge are shared across conditions. We use an exact
two-sided sign test on discordant pairs. On MemLens we deliver $K{=}10$ messages
unless stated otherwise. On DMV-Bench we keep the benchmark's injection budget of five, part of the
interface it defines.

\begin{table}[t]
\centering
\caption{Delivery against retrieval on MemLens 32K ($n=173$ non-refusal questions,
benchmark judge). Rows vary what is retrieved and columns whether the retrieved items keep
their pixels, each row paired within its own run. W/L counts questions fixed and broken,
and Appendix~\ref{app:arms} defines the conditions.}
\label{tab:decomp}
\small
\setlength{\tabcolsep}{4pt}
\begin{tabular}{lrrrl}
\toprule
Retrieved set & Pixels delivered & Pixels withheld & $\Delta_{D}$ & W/L, $p$ \\
\midrule
Gold messages ($R^{\star}$) & $43.93$ & $30.06$ & $+13.87$ & 31/7, $1.2\times10^{-4}$ \\
Our retriever, text channel ($R$) & $41.62$ & $27.17$ & $+14.45$ & 30/5, $2.2\times10^{-5}$ \\
\midrule
$\Delta_{R}$ & $+2.31$ & $+2.89$ & \multicolumn{2}{l}{interaction $-0.58$} \\
\quad W/L, $p$ & 17/13, $0.58$ & 15/10, $0.42$ & \multicolumn{2}{l}{12/14, $0.85$} \\
\bottomrule
\end{tabular}
\end{table}

Write $A(R,D)$ for accuracy under retriever $R$ and delivery $D$, $R^{\star}$ for the
oracle retriever that returns the gold messages, and $D_{0}$ for text-only delivery, the
retrieved messages passed on without their pixels. Both quantities are taken from the corner $(R^{\star}, D)$, where retrieval is perfect and the pixels are delivered,
\begin{equation}
\Delta_{D} \;=\; A(R^{\star}, D) - A(R^{\star}, D_{0}),
\qquad
\Delta_{R} \;=\; A(R^{\star}, D) - A(R, D),
\label{eq:decomp}
\end{equation}
each varying one argument and fixing the other. $\Delta_{D}$ removes the pixels and
keeps the gold messages, leaving no room for a retrieval error. $\Delta_{R}$ replaces the gold
messages with those of our text-channel retriever and keeps the pixels. Each starts from what systems do today,
text-only delivery for $\Delta_{D}$ and a query-conditioned retriever for $\Delta_{R}$, but only $\Delta_{R}$ runs all the way to a ceiling, which favours retrieval.
Table~\ref{tab:decomp} gives $\Delta_{D}=+13.87$ and $\Delta_{R}=+2.31$. Taken at our own
retriever rather than the oracle, the delivery term is $+14.45$, and the interaction between
the two axes, $-0.58$, is undetectable. Of the $31$ questions the pixels fix, $10$ fail the
way the salmon question does when the pixels are withheld, with a confident wrong answer
instead of a refusal (Figure~\ref{fig:teaser}).

A small retrieval term does not mean retrieval is solved: selecting without the query costs $27$ to $34$ points at an identical budget
(Appendix~\ref{app:iso}). Across the $173$ non-refusal questions of Table~\ref{tab:decomp}, our retriever includes at least one
gold message in the delivered ten for $96.5\%$ of questions, yet recovers on average only
$20.2\%$ of the gold messages a question needs, against $99.4\%$ for the oracle condition.
Five times the evidence coverage buys only $+2.31$. Adding our own image channel to retrieval and removing every distractor each improve accuracy by $4.05$ points, again inside the noise
(Appendix~\ref{app:coverage}). On the total headroom, the gold messages alone against retrieval over both channels, neither metric finds a significant difference at $n=173$ (SubEM, substring exact match, $-1.16$; judge $+2.31$), and even that ceiling stays far below delivering the pixels.

The delivery term depends on image content rather than on image tokens. A ladder of delivery-matched counterfactual images delivers an
identical set of image slots and varies only the content of the pixels. A real but wrong
photograph gains $+0.00$ against no photograph at all; the correct image's own
statistics with every object destroyed gain $+3.47$. Against the tightest of those controls the content-specific value
is $+11.56$ of the $+15.03$ the pixels gain in that batch (Appendix~\ref{app:placebo}). The gain is concentrated. In that batch, split by question type, delivering the correct pixels rather than none gains $+40.98$ on the $61$
questions that ask the model to read a fact off a particular image and at most $+2.86$ on
the other three (Appendix~\ref{app:concentration}). The same failure, the right evidence in context and a wrong answer, also appears where nothing is ablated. On DMV-Bench with Qwen2.5-VL-7B at $J{=}5$, where retrieval recall is $98.4$ to $100\%$, $98\%$ of our failures picked a neighbour while the correct item was in context, with the agent's own reasoning trace
naming the target correctly in $94\%$ of those. MemLens's own authors observe it in published agents that retrieve the evidence and still answer wrongly, and read it as a limit on the backbone's reasoning; the decomposition offers a different account of those errors (Appendix~\ref{app:audit}).

The delivery term could be an artefact of $K$ or of one model. Sweeping $K$ shows accuracy
climbs to ten
and stops, with $K{=}20$ scoring identically (Appendix~\ref{app:knobs}). Qwen3-VL-30B-A3B and Qwen3-VL-32B test the second, recomputing both terms of Equation~\ref{eq:decomp} on identical questions, $K$, judge and retrieval index.

Delivery remains the larger term on both larger backbones and grows with the backbone, from
$+13.87$ to $+15.61$ to $+17.92$; on the $32$B model it is the largest delivery effect we measure, $32$ questions improved against $1$ worsened (Table~\ref{tab:scale}). Retrieval grows faster, from $+2.31$ at $8$B to $+9.83$ and $+11.56$, and the ratio between them narrows from
$6.00\times$ to $1.59\times$ and then $1.55\times$. The ratio barely moves over that last step
despite a ten-fold increase in active parameters. The gap between the two terms is significant at $8$B ($p=0.0078$) and keeps its sign on the larger two without reaching significance at $n=173$ ($p=0.18$ and $0.14$, Appendix~\ref{app:scale}).

\section{Method}\label{sec:method}

\begin{figure}[t]
\centering
\includegraphics[width=\textwidth]{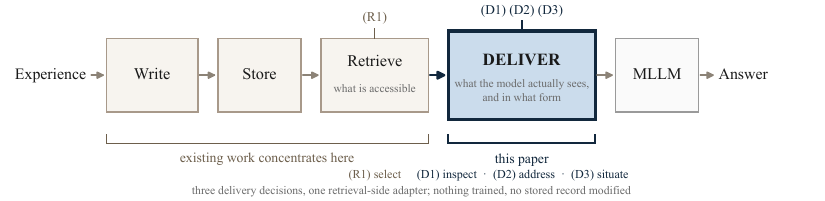}
\caption{The memory pipeline, with the stage this paper isolates. MLLM is the answering model $M$.}
\label{fig:arch}
\end{figure}

The decomposition identifies an interface failure: the relevant records reach the prompt,
but not in a form the model can act on. The failures we audited lack three properties: the model has to be able to inspect what a memory contains, to
address one item among those delivered, and to situate it in time.
DeliverMem supplies them as (D1), (D2) and (D3), at the stage Figure~\ref{fig:arch} marks. Because (D1)--(D3) train nothing
and touch only the outgoing delivered context, they retrofit any stack that keeps the
original media and their timestamps. On
DMV-Bench they are an adapter onto the benchmark's own harness, its retrieval, encoding
and store untouched.

Delivery cannot supply a fourth property: it cannot expose an item that retrieval never returned. A retrieval-side adapter, (R1), is paired with the interface to cover that gap.
(D1)--(D3) change only the form in which a fixed set of items arrives, and (R1) changes
which items arrive.

\paragraph{(D1) Original-modality preservation.}
Retrieved items are delivered with their pixels rather than as a textual proxy. In the salmon question
of Figure~\ref{fig:teaser} this is the entire difference between the two answers, because the
position is in the photograph and nowhere in the message. Table~\ref{tab:decomp} isolates this component, and Appendix~\ref{app:payload} shows it
with its controls, and Appendix~\ref{app:prompts} the full request.

\paragraph{(D2) Addressability.}
Several photographs arrive at once and counting is a poor way to point at one. Each
delivered image therefore carries a short readable tag (\texttt{M1}, \texttt{M2}, \dots)
rendered into the pixels at the top-left corner, in the manner of Set-of-Mark
prompting \citep{som}, and the prompt refers to a memory by that tag rather than by its
position. The tag is written at delivery time, after the embeddings, so addressability
costs no re-embedding and leaves the retrieval ranking untouched. Moving the tag into the text, pixels intact, costs $7.89$ points on DMV-Bench (Appendix~\ref{app:channel}).

\paragraph{(D3) Temporal position.}
A question such as ``what did I decide about this before the trip?'' cannot be answered
from a set of photographs with no times attached, however good the retrieval was. Each
delivered item is therefore annotated with its position in the memory store, using the finest temporal index the store provides: MemLens records timestamps, and
there the index is the date. DMV-Bench records none, and the index is the item's append
ordinal instead, rendered as \texttt{[seen \#$k$ of $N$]}.

\paragraph{(R1) Region multi-vector retrieval, the retrieval-side adapter.}
Images are encoded as a whole plus a $2\times2$ overlapping grid and scored by region
MaxSim \citep{colbert} over SigLIP~2 features \citep{siglip2}. It targets evidence
confined to a small region of the frame.

\paragraph{Configuration selection.} Both benchmarks use these components. A dev-split rule can also choose, per question type, between them and a baseline (Appendix~\ref{app:routing}); Tables~\ref{tab:memlens} and~\ref{tab:dmv} report the single configuration without it.

\section{Experiments}\label{sec:results}

\begin{table}[t]
\centering
\caption{MemLens at all four released context lengths (benchmark judge, canonical $195$
questions, Qwen3-VL-8B, $K{=}10$). ``Images withheld'' delivers the retrieved messages
without their pixels; ``text-channel retrieval'' keeps the pixels and drops the image
channel from retrieval. The shaded row is the one
configuration deployed on both benchmarks, applied to every question; a per-type selection
rule fitted on the dev split is reported separately (Appendix~\ref{app:routing}).}
\label{tab:memlens}
\small
\begin{tabular}{lrrrr}
\toprule
& 32K & 64K & 128K & 256K \\
\midrule
Best published agent at each length & 32.82 & 31.28 & 31.28 & 30.77 \\
\midrule
DeliverMem, images withheld & 35.38 & 35.38 & 32.82 & 30.26 \\
DeliverMem, text-channel retrieval & 48.72 & 44.10 & 44.10 & 38.46 \\
\rowcolor{black!7}
DeliverMem & 51.28 & 44.10 & 42.05 & 38.97 \\
\midrule
Margin over the best published agent & $+18.46$ & $+12.82$ & $+10.77$ & $+8.20$ \\
\midrule
Input ratio, full context to delivered & $10.0\times$ & $19.2\times$ & $36.6\times$ & $72.9\times$ \\
\bottomrule
\end{tabular}
\end{table}

\begin{table}[t]
\centering
\caption{DMV-Bench task success rate (\%) on both backbones. The upper block is the
benchmark's own pool, with NoMemory, TextOnly and Caption as its baselines and the
-lite rows as its re-implementations of published systems. The lower block is
ours, probe-pooled. The $J{=}50$ columns use the benchmark's Monte-Carlo probe
construction (Appendix~\ref{app:paired}).}
\label{tab:dmv}
\small
\setlength{\tabcolsep}{4pt}
\begin{tabular}{lrrrrcrrrr}
\toprule
& \multicolumn{4}{c}{Qwen2.5-VL-7B} & & \multicolumn{4}{c}{Gemini 2.5 Flash} \\
\cmidrule(lr){2-5} \cmidrule(lr){7-10}
Memory & $J{=}5$ & $J{=}10$ & $J{=}15$ & $J{=}50$ & & $J{=}5$ & $J{=}10$ & $J{=}15$ & $J{=}50$ \\
\midrule
NoMemory & 0.0 & 0.0 & 0.0 & 0.0 & & 0.0 & 0.0 & 0.0 & 0.0 \\
TextOnly & 0.0 & 0.0 & 0.0 & 0.0 & & 0.0 & 0.0 & 0.0 & 0.0 \\
WorldMM-lite \citep{worldmm} & 37.3 & 34.1 & 29.7 & 27.5 & & 43.5 & 41.1 & 38.2 & 37.0 \\
MMA-lite \citep{mma} & 47.7 & 41.0 & 39.4 & 35.2 & & 46.1 & 44.8 & 35.8 & 36.9 \\
Caption & 67.9 & 64.5 & 62.3 & 58.1 & & 58.9 & 55.6 & 50.5 & 47.7 \\
M2A-lite \citep{m2a} & 69.0 & 66.7 & 62.6 & 59.8 & & 65.7 & 65.2 & 61.9 & 64.7 \\
DualMem \citep{dmvbench} & 81.2 & 77.8 & 74.6 & 68.3 & & 82.7 & 75.3 & 71.3 & 65.1 \\
\midrule
\rowcolor{black!7}
DeliverMem & \textbf{88.2} & \textbf{89.7} & \textbf{88.9} & \textbf{85.0} & & \textbf{93.3} & \textbf{90.3} & \textbf{89.5} & \textbf{88.6} \\
$\Delta$ over previous best & $+7.0$ & $+11.9$ & $+14.3$ & $+16.7$ & & $+10.6$ & $+15.0$ & $+18.2$ & $+23.5$ \\
\bottomrule
\end{tabular}
\end{table}

All seven published memory agents run on $8$B parameters or fewer, the strongest on a 7B backbone against our 8B, and four read captions in place of images (Appendix~\ref{app:pool}). At $32$K the strongest, MemAgent-7B, reaches $32.82$. An $8$B model handed ten delivered messages reaches $51.28$, above the $50.77$ it scores reading the entire conversation, and our lead over the published systems persists at every setting of both benchmarks (Tables~\ref{tab:memlens} and~\ref{tab:dmv}). Of the $36$ questions behind the $32$K margin, $19$ are refusal questions, which our pipeline answers correctly even with the pixels withheld. On the other $173$ the margin is $+9.83$ points: with the pixels withheld, text-channel retrieval trails MemAgent-7B by $14$ questions, restoring the pixels gains $26$, and the image channel adds $5$ (Table~\ref{tab:pool}). The lead comes from information extraction, where the answer is in the image: we score $55.74$ there against at most $18.03$.

The benchmark's direct-LVLM pool takes the whole conversation as input
(Table~\ref{tab:lvlm}, Appendix~\ref{app:lvlm}). At $32$K our $51.28$ is level with GPT-5.4 \citep{gpt54card} on a tenth of the
input: an $8$B open model reading ten messages, against a frontier system reading
everything. Gemini-3.1-Pro \citep{gemini31pro}, Qwen3-VL-30B-A3B and Qwen3.5-122B \citep{qwen35} still score higher there ($55.38$, $58.46$ and $63.59$), and at $64$K and $128$K the frontier systems pull ahead, reading
$19$ to $37$ times what we deliver. Our delivered input barely moves as the context grows,
$2{,}405\to2{,}450$ tokens across a $7.4\times$ increase. Against direct inference with our backbone we are ahead by $0.51$ at $32$K and by $7.69$ at $128$K, and behind by
$3.59$ at $64$K
(Table~\ref{tab:cost}, Appendix~\ref{app:cost}).
Adding the tag and the date yields $+1.03$ points overall, an aggregate that
hides a split between question types (Appendix~\ref{app:subgroup}).

DMV-Bench supports two comparisons, and they use different estimates.
Table~\ref{tab:dmv} sets our score against the numbers the benchmark publishes. Both sides
are pooled over probes, the metric those numbers report, and a published figure admits no
paired test. Paired tests use a same-batch reproduction of DualMem instead. The benchmark's probes are nested inside chains, one generated per seed, so we treat the chain as the independent unit and report the mean of per-chain accuracies. The test is an exact cluster permutation \citep{permtest} over all $2^m$ sign assignments for $m$ chains, with a floor of $p=0.002$ at $m=10$. The margins below differ from Table~\ref{tab:dmv}'s because both the estimate and the opponent's run
differ.

On DMV-Bench the margin grows with $J$. Against that reproduction the paired difference is positive in every chain at every setting but Qwen at $J{=}5$, each at
the exact permutation floor for its chain count (Table~\ref{tab:dmv_paired},
Appendix~\ref{app:paired}). It increases monotonically on both backbones
from $+7.63$ to $+12.86$ on Qwen and from $+12.58$ to $+22.62$ on Gemini, while on MemLens the
analogous axis, context length, moves the other way. At $J{=}50$ its exhaustive and Monte-Carlo probe constructions give $+12.86$
and $+12.79$ (Appendix~\ref{app:paired}). This margin is the whole interface, delivery and
retrieval together. On this benchmark the retrieval decision carries most of it, and
Section~\ref{sec:controls} separates the two inside a $2\times2$; the claim that delivery is the
larger term is a statement about MemLens alone.

\section{Attribution and Transfer under Delivery-Matched Controls}\label{sec:controls}\label{sec:isobudget}\label{sec:boundary}

Every delivery component below is paired with a control $D_{c}$ that agrees with the
delivery $D$ on the delivered items and their order, and differs in the one variable that component sets. Without that, a mechanism can appear to
work while the effect is entirely attributable to how its output is rendered
\citep{presentation}.
We report $\Delta = A(R, D) - A(R, D_{c})$, with $R$ our retriever or, for (D1), $R^{\star}$. An ablation, by contrast, moves several coordinates of the
delivered context at once. Table~\ref{tab:controls} collects the results.

\begin{table}[t]
\centering
\caption{Each delivery component against a delivery-matched control, and (R1) inside a $2\times2$. Agreement is wins/losses over discordant questions on MemLens and chains agreeing in sign on DMV-Bench. MemLens rows are at 32K, the (D3) row on the temporal questions of all $789$ released; DMV rows use Qwen2.5-VL-7B at $J{=}10$, (R1) at $J{=}50$. The first three DMV rows nest, each control removing more than the one above. $^{*}$Against a constant tag rather than no tag.}
\label{tab:controls}
\small
\setlength{\tabcolsep}{3.5pt}
\begin{tabular}{llllrl}
\toprule
Component & Control alters & Benchmark & Unit ($n$) & $\Delta$ & Agreement, $p$ \\
\midrule
(D1) original modality & pixel phase randomised & MemLens & question (173) & $+11.56$ & 28/8, $0.0012$ \\
(D2) addressability & rank$\to$tag map shuffled & DMV & chain (10) & $+2.63$ & 8 of 10, $0.072$ \\
(D2) addressability$^{*}$ & tags made identical & DMV & chain (10) & $+11.85$ & 10 of 10, $0.002$ \\
(D2) addressability & tag removed entirely & DMV & chain (10) & $+9.88$ & 10 of 10, $0.002$ \\
(D2) addressability & tag in text, pixels intact & DMV & chain (10) & $+7.89$ & 9 of 10, $0.0039$ \\
(D2) addressability$^{*}$ & tags made identical & MemLens & question (195) & $+0.51$ & 4/3, $1.00$ \\
(D3) temporal position & date values permuted & MemLens & question (194) & $+13.92$ & 31/4, $3.5\times10^{-6}$ \\
Binding & image$\to$message shuffled & MemLens & question (173) & $+6.36$ & 14/3, $0.013$ \\
\midrule
(R1) region retrieval & (main effect in $2\times2$) & DMV & chain (10) & $+8.83$ & 10 of 10, $0.002$ \\
\bottomrule
\end{tabular}
\end{table}

\begin{table}[t]
\centering
\caption{Component by benchmark, with the contrast behind each entry given in the text. On DMV-Bench, (D2) and (R1) are main effects in the $2\times2$ design with Qwen2.5-VL-7B at $J{=}50$, whose delivery condition moves the tag and the ordinal together, and (D2) is null with Gemini~2.5~Flash; $\dagger$~marks (D3) against no date, on temporal-reasoning questions. A dash marks a component the benchmark cannot separate or did not measure.}
\label{tab:matrix}
\small
\begin{tabular}{lll}
\toprule
Component & DMV-Bench & MemLens \\
\midrule
\multicolumn{3}{l}{Delivery} \\
\quad (D1) original modality & --- & positive $+11.56$ \\
\quad (D2) addressability & positive $+3.96$ & null \\
\quad (D3) temporal position & null & positive $+8.76^{\dagger}$ \\
\quad Binding (not deployed) & --- & positive $+6.36$ \\
\midrule
\multicolumn{3}{l}{Retrieval} \\
\quad (R1) region retrieval & positive $+8.83$ & null \\
\bottomrule
\end{tabular}
\end{table}

Table~\ref{tab:matrix} condenses this section's results by benchmark, and every null in it is one of the three conditions of Section~\ref{sec:intro} failing. A
DMV-Bench question asks which of several retrieved frames contains the cue. A name for each frame
answers it directly, and an ordinal adds nothing. A MemLens temporal question asks when something
happened: it needs the date, and a name for each item does not supply one. The split by question type in Section~\ref{sec:decomp} follows the rule inside one benchmark.

The (D2) controls each remove something different. Shuffling the tags removes the correct tag-to-item mapping, and stamping one symbol on every image removes the name itself. Moving the tag into the text (Section~\ref{sec:method}) takes the name out of the pixels; removing it leaves no marker at all; and the $2\times2$ below removes the tag and the ordinal together. Removing the mapping has a small effect that is not significant, because the delivered text is already enumerated and position leaks rank (Appendix~\ref{app:arms}). Removing the name matters. Against the constant symbol the deployed tag gains $11.85$ points on DMV-Bench, all ten chains agreeing in sign, and shuffled tags, unique but attached to the wrong items, still gain $9.23$ over the constant symbol, which isolates uniqueness, the property (D2) supplies. Against no tag at all, with the ordinal, item text, image set and order held, the gain is $9.88$, again on all ten chains at the permutation floor (Appendix~\ref{app:channel}). The constant symbol thus sits $1.97$ points below no marker ($p=0.078$), so the marker's own cost is small.

On MemLens that contrast is half a point, though replacing every tag with one symbol still changes the answer on $27$ of $195$ questions, and the seven that change the score split four gains against three losses.

We test (D3) for specificity with a single interaction over the $699$ non-refusal questions of the $789$ released. Accurate dates, set
against a value-preserving permutation of them, gain $+13.92$ on the $194$
temporal-reasoning questions while the other $505$ move by $-0.40$;
the difference-in-differences, tested by permuting the group label, is $+14.31$. The system-level
net over all $789$ is $+3.17$ ($37$/$12$, $p=4.7\times10^{-4}$), under a quarter of the
effect on temporal questions, as expected once an effect confined to one question
type is averaged over every question. This shape recurs at all four context lengths (Appendix~\ref{app:d3len}).
A third condition with no date at all splits the gain. In a run that includes all three, again on temporal questions, accurate dates improve on no date by $8.76$ points and on permuted dates by $14.43$, one question from the $+13.92$ above. The $5.67$-point difference is the harm an incorrect date does (Appendix~\ref{app:d3len}).

Delivered images could help merely by adding salience. A binding control tests that by permuting only which image follows which message (Appendix~\ref{app:arms}): accuracy falls by $6.36$ points at identical gold coverage and image count, so the pairing of image and message matters in its own right.

The retrieval-side and
delivery-side components are not independent. At $J{=}50$ on DMV-Bench with
Qwen2.5-VL-7B, against the
same-batch DualMem run, delivery alone improves on that run by $1.39$ points, (R1) alone by $6.27$ and the two
together by $12.79$. That $2\times2$ gives (R1) a main effect of $+8.83$ and delivery, entered under
(D2) in Table~\ref{tab:matrix}, a main effect of $+3.96$ ($p=0.0039$), with a significant interaction of
$+5.14$ (8 of 10 chains, $p=0.020$). (R1) and delivery are therefore reported jointly rather
than as independently additive contributions. This design also separates (R1)'s encoder from delivery, which otherwise confounds the pair at the system level \citep{memdelta}. On Gemini~2.5~Flash these four cells put nearly the whole margin on (R1) (Appendix~\ref{app:gemini2x2}).

An effect tied to one benchmark would be expected from (R1), since it is the only
component that changes which items arrive. It is positive on DMV-Bench on both backbones and null on MemLens, where replacing the default index with the region index moves the same condition by $-1.03$ ($4$/$6$, $p=0.75$, paired across two runs; Table~\ref{tab:pool}).
Its target is a small image region, and only DMV-Bench is built that way: its cue occupies a median $1.7\%$ of the frame (Appendix~\ref{app:region}). The delivery components split across the benchmarks as well, so whether a component transfers does not follow from the stage it acts on. (D2) is positive on DMV-Bench with Qwen2.5-VL-7B and null on MemLens, whose questions need a date or a fact more than a name (condition~(ii)). The second backbone makes condition~(i) visible, since the item has to arrive before a name for it can help, and the condition predicts delivery as a whole across the settings we measured: delivery is the larger term on MemLens, where a gold message reaches the
model on $96.5\%$ of questions; it has a positive main effect on DMV-Bench with Qwen2.5-VL-7B,
where the baseline's failures are selections among items already in context; and it is null on
Gemini~2.5~Flash, where $92\%$ of the baseline's failures are retrieval misses (Appendix~\ref{app:audit}). (D3) is
the mirror image, positive on MemLens temporal questions and null on DMV-Bench. There, adding
the date on top of the tag moves the score by $+0.07$ at $J{=}10$ with Qwen2.5-VL-7B, $29$ of $800$ paired probes gained and $27$ lost. (D1) cannot be
measured on DMV-Bench, whose baseline already delivers pixels.

One assembled configuration nonetheless leads on both benchmarks, applied to every question
with no selection at all (Tables~\ref{tab:memlens} and~\ref{tab:dmv}). Of our components, delivery does the work on MemLens and (R1) most of it on DMV-Bench.

\section{Limitations}

(1) Delivery granularity, whether to send a crop, the whole frame, or both, is a
degree of freedom we do not measure. MemLens photographs have no well-defined relevant
region, and on DMV-Bench the axis is already occupied on the retrieval side by (R1), with
which it interacts. Separating them calls for a benchmark with region annotations,
which we leave to future work.
(2) Per-type effects are measured on the full $789$ released questions, since the $195$-question
evaluation subset contains only $29$ knowledge-update questions (Appendix~\ref{app:subgroup}).
(3) Every measurement is on the image modality (Appendix~\ref{app:modality}).

\section{Conclusion}

The salmon question of Section~\ref{sec:intro} was answered wrongly with the right message in
the prompt, and correctly once its pixels arrived. Deciding what reaches the model, and in what form, is a stage of its own between retrieving a memory and answering from it. Starting from today's
text-only delivery and query-conditioned retrieval, that stage has more accuracy left in it than
retrieval, significantly on Qwen3-VL-8B and in the same direction on the two larger ones. Making its three decisions explicit, with nothing trained,
turns a pipeline that trails the strongest published agent on non-refusal questions into one that leads it; with the retrieval adapter, one configuration leads both benchmarks. The three decisions are a
first pass at the stage: binding, measured at $+6.36$, stays outside the interface,
and audio and video will bring decisions of their own.

\subsection*{AI use statement}

The systems this paper studies contain large models of their own: the MemLens evaluation judge, and the captioner behind DMV-Bench's Caption baseline. Both are parts of the
benchmarks' own pipelines.
We also used generative AI tools for polishing and editing the text of this paper, for
writing the experiment and analysis code, for running and monitoring the experiments, and
for literature search. Generative AI was not used to generate, augment or modify any experimental data. We have
reviewed all AI-assisted work: every accuracy figure we measured ourselves is recomputed from the stored per-question judge scores and per-probe outcomes, and figures taken from other papers are transcribed from them. We take
responsibility for the final content of this work, including text, claims and artifacts
produced with the aid of generative AI.

\section*{Ethics Statement}

This work involves no human subjects and collects no new data. Both benchmarks are
public and are used as released. Their images are their own, and DMV-Bench's cued
frames are synthetic edits of studio photographs rather than pictures of people. 

The method has one consequence for privacy. (D1) exists because delivering
the stored pixels beats delivering a textual proxy, so a memory system built this way
places more raw imagery in front of the answering model than a captioning system does.
Where the stored media are personal, that widens what the model sees, and a deployment
would need to decide separately what may be retained and re-shown. The interface itself
neither writes to the store nor alters it: (D2) stamps a tag into the copy that is
delivered, and the stored record is untouched.

\section*{Reproducibility Statement}

Code will be released on publication, together with the per-question and per-probe outputs behind our tables: the judge score of each MemLens question and the outcome of each DMV-Bench probe under every condition our tables report, and the agent traces behind the worked failure cases. Every paired contrast in our tables can be recomputed from those outputs without re-running a model. Tables reporting other systems' published numbers are transcribed from their sources
and marked as such. The conditions behind every contrast are
defined in Appendix~\ref{app:arms}, and those of the counterfactual-image ladder in
Appendix~\ref{app:placebo}. The paired sign test used on MemLens and the exact
cluster permutation used on DMV-Bench are stated in Sections~\ref{sec:decomp}
and~\ref{sec:results}, with the variance decomposition behind the latter in
Appendix~\ref{app:variance}.

Experiments ran on a single $32$\,GB GPU under Python~$3.10$ with PyTorch~$2.11$ and Transformers~$5.15$. Qwen3-VL-8B on MemLens and Qwen2.5-VL-7B on DMV-Bench are served locally behind an OpenAI-compatible endpoint; Gemini~2.5~Flash, the two larger backbones of Table~\ref{tab:scale} and the MemLens judge, Qwen3-VL-235B-A22B, are called through hosted APIs behind that same interface. DMV-Bench is scored exactly, by whether the agent opens the correct product page. Images are encoded with \texttt{siglip2-base-patch16-384} and the retriever's text channel with \texttt{bge-small-en-v1.5} \citep{cpack}. Re-running either benchmark end to end additionally requires that benchmark's
data and harness, which we do not redistribute.

\bibliographystyle{iclr2027_conference}
\bibliography{refs}

\clearpage
\appendix

\section{Benchmarks, Setup and Comparison Pools}

\subsection{Benchmark Setup}\label{app:setup}

DMV-Bench \citep{dmvbench} was introduced as the first interactive benchmark for an
agent's visual memory, on the observation that existing benchmarks ask what an
agent could write down rather than what it had to see. The cue composited into one product photograph comes from a non-repeating vocabulary of colour--object pairs, and a recall probe names the cue and asks the agent to navigate back to its product; the metric is task success rate, an exact match of the product page. Its three reference baselines keep no memory (NoMemory), index each page's bare product class (TextOnly), or index a generated caption of each product image (Caption). The probes are nested, and the benchmark's authors note that treating them as i.i.d.\
inflates apparent significance; our statistical protocol is built on that point. The dual-bank method, DualMem, runs at the authors' best reported fusion weight $\alpha{=}0.75$, $3.0$
points above the symmetric default in their own sweep. DMV-Bench ships no caption cache. We rebuilt one with Gemini~2.5~Flash and the original prompt, and our captions name the cue in $16.1\%$ of cases, against
the $16.5\%$ the benchmark reports for its own.

In MemLens's answer-refusal questions \citep{memlens}, the correct response is to say that the context does not contain the answer.

\subsection{Where DeliverMem Acts in Each Benchmark}\label{app:placement}

Table~\ref{tab:placement} follows one query through both pipelines, from the store to the score.
DeliverMem changes the two shaded rows and nothing else. On DMV-Bench every unshaded row is the
benchmark's own DualMem configuration, and our delivery replaces only the step that assembles what the model receives
(Appendix~\ref{app:prompts}). The unshaded MemLens retrieval rows belong to the baseline we built
for this paper.

\begin{table}[ht]
\centering
\small
\caption{Where DeliverMem acts in each benchmark, in pipeline order. Shaded rows are the stages
DeliverMem changes.}
\label{tab:placement}
\renewcommand{\arraystretch}{1.22}
\begin{tabular}{@{}>{\raggedright\arraybackslash}p{0.15\linewidth}>{\raggedright\arraybackslash}p{0.40\linewidth}>{\raggedright\arraybackslash}p{0.39\linewidth}@{}}
\toprule
Stage & MemLens & DMV-Bench \\
\midrule
Store & every conversation message with its images, $32$K to $256$K tokens per conversation
      & every product page visited over $J$ sessions, in DualMem's image and text banks \\
Text channel & \texttt{bge-small-en-v1.5} \citep{cpack} on the message text & \texttt{all-MiniLM-L6-v2} \citep{sbert} on Gemini~2.5~Flash captions \\
\rowcolor{black!7}
Image channel (R1) & \texttt{siglip2-base-patch16-384}, whole image plus $2\times2$ overlapping
      tiles, scored by MaxSim
      & the same encoder and tiles, in place of DualMem's whole-image vector \\
Fusion & image scores min-max normalised within the question and averaged with the text score;
      a message without an image scores $0.5$ on the image channel
      & DualMem's weighted fusion at $\alpha{=}0.75$ \\
Items delivered & $K{=}10$ messages & the five items the benchmark specifies \\
\rowcolor{black!7}
Delivery (D1--D3) & original images (D1), a tag in the pixels (D2) and the message date (D3)
      & a tag in the pixels (D2) and \texttt{[seen \#$k$ of $N$]} (D3); DualMem already delivers images \\
Answering model & Qwen3-VL-8B; Qwen3-VL-30B-A3B and Qwen3-VL-32B in Table~\ref{tab:scale}
      & Qwen2.5-VL-7B or Gemini~2.5~Flash, acting in the storefront \\
Scoring & the benchmark's judge, Qwen3-VL-235B-A22B
      & task success: the agent opens the exact product page \\
\bottomrule
\end{tabular}
\end{table}

\subsection{Choosing the Reference System}\label{app:refsys}

Table~\ref{tab:dmv} and Table~\ref{tab:pool} (Appendix~\ref{app:pool}) give the published
pools in full. The three -lite rows of Table~\ref{tab:dmv} are the benchmark's own
re-implementations of published systems, since the original systems do not run against DMV-Bench as released. DualMem is the strongest published system in all eight DMV-Bench cells, and
NoMemory and TextOnly score $0.0$ throughout. On MemLens the strongest of the seven
published memory agents reaches $32.82$, and the ordering among them shifts across question types: MemAgent-7B is best overall yet scores $13.64$ on refusal, while
Mem0 scores $77.27$ there and much lower elsewhere. Per-type results are therefore given in Table~\ref{tab:pool},
but the reference point is the best system overall.
Direct inference with our own backbone reaches $50.77$. Because it reads the entire context, it belongs to the long-context family and serves
as a cost--accuracy reference
(Appendix~\ref{app:cost}).

\subsection{MemLens 32K by Question Type}\label{app:pool}

Table~\ref{tab:pool} gives the published pool together with the direct-LVLM pool
and our conditions, broken out by the benchmark's five question types. Every one of the seven published memory agents runs on a backbone of $8$B parameters or
fewer: Qwen3-VL-8B (M2A), Qwen3-8B (Mem0, MemOS), Qwen2-VL-7B (M3-Agent), Qwen2.5-7B
(MemAgent-7B), Qwen2.5-3B (Memory-T1) and Qwen2-VL-2B (M3C). Four of them are text-only and receive BLIP-2 \citep{blip2} captions in place of every evidence image. In four published rows the
per-type entries do not sum to the overall accuracy given in their source, and we reproduce both as published. Our own rows come from two runs, the upper block on the default image index, one whole-image vector per image, and the lower on the region index of the deployed configuration. At the other three lengths the best published agent is Mem0 at 64K and 256K and MemOS at 128K \citep{memlens}.

\begin{table}[t]
\centering
\caption{MemLens at 32K by question type (canonical $195$ questions, benchmark judge),
with every published memory agent in the benchmark's pool. IE, MSR, TR, KU and AR are information extraction, multi-session reasoning, temporal reasoning, knowledge update and answer refusal. Published rows are transcribed as reported, and the shaded row is the configuration of Table~\ref{tab:memlens}.}
\label{tab:pool}
\small
\setlength{\tabcolsep}{3.5pt}
\begin{tabular}{lrrrrrrr}
\toprule
& all & non-AR & & & & & \\
System & (195) & (173) & IE & MSR & TR & KU & AR \\
\midrule
M2A \citep{m2a} (same backbone) & 15.38 & 14.45 & 14.75 & 8.57 & 2.08 & 0.00 & 22.73 \\
M3C \citep{m3c} & 18.46 & 18.50 & 8.20 & 25.71 & 31.25 & 10.34 & 18.18 \\
M3-Agent \citep{m3agent} & 19.49 & 20.23 & 18.03 & 22.86 & 29.17 & 6.90 & 13.64 \\
Memory-T1 \citep{memoryt1} & 28.72 & 31.21 & 18.03 & 25.71 & 62.50 & 13.79 & 9.09 \\
MemOS \citep{memos} & 30.26 & 25.44 & 18.03 & 22.86 & 39.58 & 24.14 & 68.18 \\
Mem0 \citep{mem0} & 31.79 & 26.01 & 13.11 & 25.71 & 50.00 & 17.24 & 77.27 \\
MemAgent-7B \citep{memagent} & 32.82 & 35.26 & 18.03 & 25.71 & 62.50 & 41.38 & 13.64 \\
\midrule
\multicolumn{8}{l}{Direct LVLMs, reading the entire context \citep{memlens}} \\
Qwen3.5-122B \citep{qwen35} & 63.59 & 59.54 & 75.41 & 45.71 & 58.33 & 44.83 & 95.45 \\
Qwen3-VL-30B-A3B \citep{qwen3vl} & 58.46 & 53.76 & 65.57 & 25.71 & 62.50 & 48.28 & 95.45 \\
Gemini-3.1-Pro \citep{gemini31pro} & 55.38 & 49.71 & 60.66 & 45.71 & 39.58 & 48.28 & 100.00 \\
GPT-5.4 \citep{gpt54card} & 51.28 & 45.09 & 63.93 & 17.14 & 35.42 & 55.17 & 100.00 \\
Qwen3-VL-8B \citep{qwen3vl} (our backbone) & 50.77 & 46.82 & 52.46 & 22.86 & 58.33 & 44.83 & 81.82 \\
Qwen3-VL-2B \citep{qwen3vl} & 38.46 & 35.26 & 27.87 & 20.00 & 62.50 & 24.14 & 63.64 \\
\midrule
\multicolumn{8}{l}{Ours, default image index} \\
\quad images withheld & 35.38 & 27.17 & 11.48 & 22.86 & 39.58 & 44.83 & 100.00 \\
\quad text-channel retrieval & 48.72 & 42.20 & 55.74 & 17.14 & 39.58 & 48.28 & 100.00 \\
\quad text and image channels & 51.28 & 45.09 & 57.38 & 22.86 & 37.50 & 58.62 & 100.00 \\
\midrule
\multicolumn{8}{l}{Ours, region multi-vector index (a separate run)} \\
\quad without tag or date & 50.26 & 43.93 & 59.02 & 17.14 & 39.58 & 51.72 & 100.00 \\
\rowcolor{black!7}
\quad DeliverMem, with tag and date & 51.28 & 45.09 & 55.74 & 20.00 & 54.17 & 37.93 & 100.00 \\
\bottomrule
\end{tabular}
\end{table}

\subsection{Direct-LVLM Pool on MemLens}\label{app:lvlm}

Table~\ref{tab:lvlm} lists the benchmark's second method family in full. These models take the whole conversation as input, at more than ten times the inference cost, and the benchmark evaluates them only up to
$128$K.

\begin{table}[ht]
\centering\small
\caption{Direct LVLMs reading the entire MemLens context, as reported by the benchmark
\citep{memlens}, with our own row from Table~\ref{tab:memlens}. Bold and underline mark the best and second-best value in each of the first three columns.}
\label{tab:lvlm}
\begin{tabular}{lrrrr}
\toprule
& 32K & 64K & 128K & 256K \\
\midrule
Qwen3.5-122B \citep{qwen35} & \textbf{63.59} & \textbf{59.49} & 49.23 & n/a \\
Qwen3-VL-30B-A3B \citep{qwen3vl} & \underline{58.46} & 52.31 & 50.77 & n/a \\
Gemini-3.1-Pro \citep{gemini31pro} & 55.38 & \underline{58.46} & \textbf{58.46} & n/a \\
GPT-5.4 \citep{gpt54card} & 51.28 & 56.92 & \underline{51.79} & n/a \\
Qwen3-VL-8B \citep{qwen3vl} (our backbone) & 50.77 & 47.69 & 34.36 & n/a \\
Qwen3-VL-2B \citep{qwen3vl} & 38.46 & 35.38 & 27.69 & n/a \\
\midrule
\rowcolor{black!7}
DeliverMem & 51.28 & 44.10 & 42.05 & 38.97 \\
\bottomrule
\end{tabular}
\end{table}

\subsection{Comparison with Published Numbers}\label{app:published}

Table~\ref{tab:published} gives the probe-pooled comparison of Section~\ref{sec:results} row
by row, with the number of probes behind each. At $J{=}5$ we lead the published number by
$+7.02$ and the same-batch run of Table~\ref{tab:dmv_paired} by $+7.63$. The sampled $J{=}50$ row pools about a tenth of the
probes of the exhaustive row and estimates the same quantity.

\begin{table}[ht]
\centering\small
\caption{DeliverMem against the numbers DMV-Bench reports for DualMem
\citep{dmvbench}, the strongest system in its published pool. Scores are probe-pooled task
success rate (\%), $n_r$ is the number of probes pooled, and $\Delta$ is DeliverMem minus
published. The settings marked exh.\ and MC use the benchmark's exhaustive and Monte-Carlo
probe constructions.}
\label{tab:published}
\begin{tabular}{llrrrr}
\toprule
Backbone & Setting & DeliverMem & $n_r$ & Published & $\Delta$ \\
\midrule
\multirow{5}{*}{Qwen2.5-VL-7B}
 & $J{=}5$ & 88.22 & 433 & 81.2 & $+7.02$ \\
 & $J{=}10$ & 89.67 & 1{,}888 & 77.8 & $+11.87$ \\
 & $J{=}15$ & 88.87 & 4{,}359 & 74.6 & $+14.27$ \\
 & $J{=}50$ (exh.) & 85.19 & 48{,}701 & 68.3 & $+16.89$ \\
 & $J{=}50$ (MC)   & 84.96 & 4{,}820 & 68.3 & $+16.66$ \\
\midrule
\multirow{4}{*}{Gemini 2.5 Flash}
 & $J{=}5$       & 93.29 & 686  & 82.7 & $+10.59$ \\
 & $J{=}10$ & 90.32 & 3{,}079 & 75.3 & $+15.02$ \\
 & $J{=}15$      & 89.46 & 7{,}164 & 71.3 & $+18.16$ \\
 & $J{=}50$ (MC) & 88.61 & 2{,}940 & 65.1 & $+23.51$ \\
\bottomrule
\end{tabular}
\end{table}

\section{Extended Related Work}

\subsection{The Post-Retrieval Stage in Prior Work}\label{app:related}

Recent text-only work names the stage after retrieval as a ``distinct reliability boundary between retrieval and answer generation'' \citep{postretrieval} and as a
``Memory-Action Gap'' for information already in the prompt yet unused
\citep{memarbiter}. Both are text-only agents, where delivery never has to decide
modality. Presentation is itself a confound in evaluating any of this: a mechanism
reporting $+0.182$ retains $+0.021$ once a render-matched control equalises layout
\citep{presentation}, and changing only the embedding model can reorder systems
\citep{memdelta}. Both findings are why Section~\ref{sec:controls} uses delivery-matched conditions and a
$2\times2$ that keeps the encoder inside the design. The closest multimodal systems change retrieval instead: one escalates to raw sources when the query is modality-biased \citep{mmexam}, and another routes retrieval by modality \citep{vmem}. \citet{memharness} attribute failures to what happens after retrieval and train
a policy to reconstruct memory before acting; we measure that stage directly and change it without training. \citet{diagnosing} identify retrieval as the bottleneck on text-only LoCoMo
\citep{locomo}, a setting with no modality transformation between the two stages.

The stage this paper isolates lies
between a successful retrieval and the answer. A systems
characterisation calls it ``prompt assembly'' and defines it as the interface that
yields ``the prefill length of the final answer-generation call'' \citep{agentmemsys}.
The characterisation is a cost model: it prices the stage and leaves its effect on accuracy unmeasured, and its treatment of the multimodal case anticipates pressure on construction, storage and retrieval without reaching assembly.

The rule of Section~\ref{sec:intro} parallels fuzzy-trace theory \citep{ftt}, which
treats a caption as the gist of an image and the pixels as its verbatim trace;
on that account which of the two a judgment needs depends on the judgment rather than on the memory. Section~\ref{sec:controls} arrives at this pattern from data.

\subsection{Where Pixels Survive}\label{app:unit}

Whether pixels survive until delivery does not follow the obvious split between embodied
and conversational agents. Embodied agents differ among themselves: ReMEmbR
\citep{remembr} navigates from a store containing only ``the vector representation of the
text captions, the position, and the timestamps.'' The axis that separates systems is the
unit of memory. A system whose unit is a discrete frame tends to retain and deliver
that frame; one whose unit is a temporal segment or a dialogue turn summarises it at write
time, and the pixels are gone before retrieval is ever consulted.

\section{Method and Control Construction}

\subsection{Control Construction}\label{app:arms}

Each condition below changes one aspect of the delivered context relative to its reference.

Table~\ref{tab:decomp} uses four conditions and Section~\ref{sec:decomp} two more; they differ in what is retrieved and in whether the retrieved items keep their pixels. \texttt{cap10} retrieves
over the text channel. \texttt{oracle10} replaces the retrieved set with the gold messages in conversation order, cut to the first $K$ or padded to $K$ with the text channel's highest-ranked other messages, and retrieval is then at its ceiling for that $K$. \texttt{oracle10\_noimg} and
\texttt{cap10\_noimg} are those two with the pixels removed and the message set
identical. Together they form the pixels-removed column of Table~\ref{tab:decomp}.
\texttt{goldonly} delivers the gold messages with no distractors, and \texttt{img10} adds our own image channel to the text channel of \texttt{cap10} and fuses the two scores (Table~\ref{tab:placement}). 

The shuffled-tag control keeps the tag set $\{\texttt{M}1,\dots,\texttt{M}n\}$, the stamp
positions, the date fields and the instruction text verbatim, and permutes only
which tag number goes on which image. Because the delivered text also numbers the items \texttt{(1) (2) (3)} in rank order, an
item's position still gives away its rank after the shuffle. The shuffle therefore tests the tag-to-item map, with every
image still named; the constant-tag control tests the name itself, stamping the same two-character symbol on every delivered image while
holding the tag geometry, dates, item text and note syntax fixed, and replacing the
instruction sentences that promise per-image tags, two on DMV-Bench and all three on MemLens (Appendix~\ref{app:prompts}). Left in place, those sentences would
contradict the images, whereas the condition is meant to measure an absence.

The binding control fixes the stamps, dates, text, order and image set and permutes only which image follows which message.

\subsection{Scope of the Modality Claim}\label{app:modality}

Every number in this paper is measured with images as the original modality. The
argument is not specific to pixels. A transcript discards prosody, speaker identity and
overlapping speech in the way a caption discards pixels, and the decomposition is available wherever a memory system renders one modality into another before the
answering model sees it.

Temporal position (D3) acquires a second axis. On images it answers ``when was this
recorded''; on a recording it must also answer ``where within it'', and a question such
as ``what was said just after an event'' needs the two aligned rather than merely both
present. Granularity becomes a decision of its own: a MemLens photograph has no well-defined
relevant region, but an hour of audio cannot be delivered whole: the span must be chosen on every query, and that choice is a delivery
decision with no counterpart in the image case. Referring to a delivered item (D2) and preserving the
original (D1) transfer unchanged, and the binding control of
Appendix~\ref{app:arms} has a direct analogue in whether the model treats the audio and
the video of one event as one event.

\subsection{Which Channel Carries the Identity}\label{app:channel}

(D2) writes the tag into the delivered pixels. The alternative writes it into the
text that introduces the item and leaves the image untouched. The text version costs
nothing in occlusion and places no limit on how long an identifier may be. We measure the
difference directly.

\paragraph{The two variants.} Both start from the deployed configuration and differ in one place.
The deployed variant delivers \texttt{(i) [seen \#k of N] <text> [tagged M1 in its top-left
corner]} with the tag stamped into the image; the control delivers \texttt{(i) [seen \#k of
N] [M1] <text>} with the image byte-identical to the stored one. Items, order, image set,
ordinals and the tag-numbering rule are identical, and with them the tag-to-item map. The instruction sentence that says where the tag is changes, because that is the
fact under test.

Besides where the identity lives, the two variants differ in two small ways: the stamped variant's per-item note is longer than the text variant's bracketed tag, and the stamp covers a corner of the image that the text variant leaves intact.

\paragraph{Result.} We use the ten chains at $J{=}10$ behind every (D2) contrast and score both variants on $1{,}888$ shared probes. Stamping wins by $7.89$ points, nine of the ten chains agreeing in sign,
$p=0.0039$ under the exact cluster permutation used elsewhere. Repeating it at $J{=}5$ on those ten chains gives $+8.73$ (eight of ten, $p=0.0156$).

\paragraph{Neither channel works alone.} The contrast above moves the identity from the
pixels to the text. Two further conditions remove it from one channel without putting it in
the other: one stamps $\texttt{M}1\dots\texttt{M}n$ into the pixels and says
nothing about them, and the other stamps nothing. This pair is exactly
token-matched, its delivered text byte-identical.
With the deployed variant and the text-only variant it completes a $2\times2$ over those $1{,}888$ probes (Table~\ref{tab:channel2x2}).

\begin{table}[ht]
\centering\small
\caption{Where the identity lives, on DMV-Bench at $J{=}10$ (task success rate \%,
seed-averaged over ten chains). Columns are whether the delivered text names the
tag, rows whether the pixels carry it.}
\label{tab:channel2x2}
\begin{tabular}{lrr}
\toprule
& text says nothing & text names the tag \\
\midrule
pixels unstamped & $79.67$ & $82.75$ \\
pixels stamped & $81.00$ & $90.64$ \\
\bottomrule
\end{tabular}
\end{table}

Neither channel alone recovers much. Against delivering neither, the stamp with nothing
said about it is $+1.33$ ($p=0.48$, six of ten chains
agreeing in sign); the sentence with no stamp, $+3.08$ ($p=0.0059$, nine of ten); the two
together, $+10.97$ ($p=0.0039$, nine of ten), an interaction of $+6.56$ that exceeds either single-channel effect. The two columns come from different batches (Appendix~\ref{app:batches}); within one batch, delivering both against neither is the $+9.88$ of Table~\ref{tab:controls}. The identity is carried neither by the pixels nor by the text but by their agreement,
which is why (D2) is stated as a tag in the pixels and an instruction that names it.

The pattern is not a property of one working point. Repeating all four cells at $J{=}5$
(ten chains, $433$ probes) reproduces it: delivering the identity is $+13.88$ against delivering
none, every chain agreeing in sign, and the interaction stays positive at $+4.81$ against $+6.56$
at $J{=}10$. At neither working point does either channel alone reach what the two reach together.

\paragraph{Does the stamp cover the cue?} Covering the cue could only hurt the stamped variant, which
makes the $+7.89$ conservative, and it is rare. Because the stamp occupies a fixed
corner, the relevant quantity is how often cues fall in that corner. The region that wins
MaxSim serves as a proxy for where the cue is, since it contains the cue in almost every
isolated-cue frame (Appendix~\ref{app:region}). Across $76{,}927$ logged selections the
top-left tile wins $20.4\%$ of the time, below the $25\%$ a uniform cue would give, and the
stamp covers about a tenth of that tile's area. Together the two discounts leave roughly
two percent of frames whose cue the stamp could touch.

\subsection{The Delivered Context}\label{app:payload}

\begin{figure}[t]
\centering
\includegraphics[width=\linewidth]{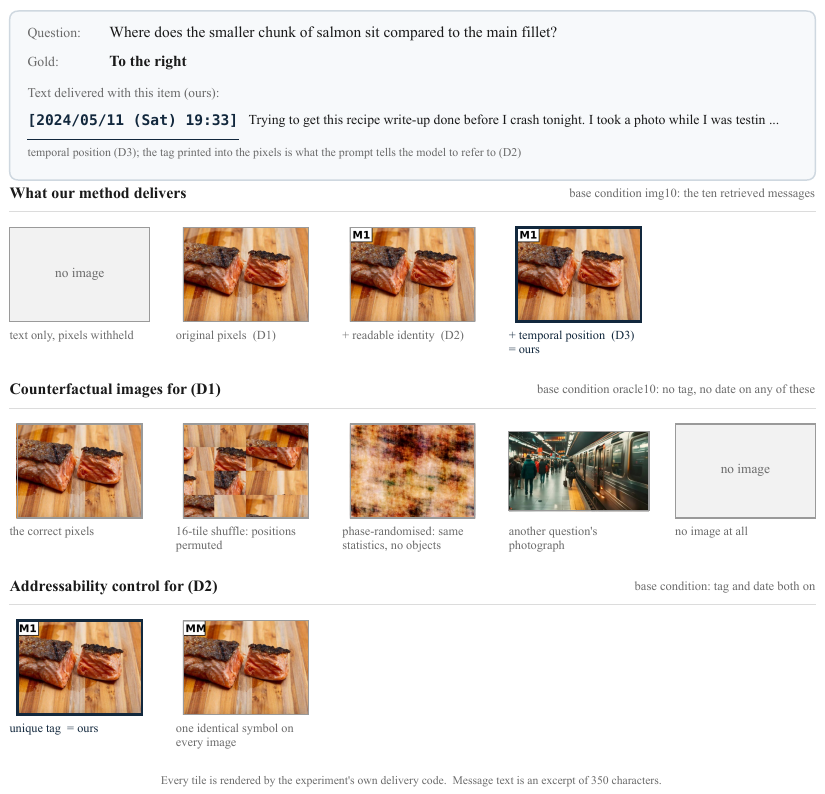}
\caption{One retrieved MemLens memory delivered eleven ways. The top block is its
accompanying text, with the (D3) timestamp in brackets and the (D2) tag in the pixels. Each
row group is a separate experiment with its base condition named at the right, and tiles
compare only within a group.}
\label{fig:payload}
\end{figure}

Every control in this paper is specified as a delivered context: a condition and its
control differ in bytes, and Figure~\ref{fig:payload} shows those bytes for one question.

\subsection{Configuration Selection}\label{app:routing}

Both benchmarks use one rule: take the configuration that is better overall on
that benchmark's dev split, and switch on a subset only when its dev effect exceeds three points. The two are the full configuration and a baseline that retrieves over both channels on the default image index and delivers the images without the tag or the date. On MemLens the full configuration is the better overall on dev,
$44.44$ against $42.09$, and is therefore the default. Knowledge update is the only subset
where the baseline is better by more than the floor, by $10.34$ points over $87$ dev
questions, and the remaining four fall inside it. DMV-Bench has a single
question class, and there the rule reduces to the globally better configuration. On a
held-out seed that configuration wins at all four settings, by $11.11$, $8.61$, $7.77$ and
$5.79$ points. The rule is fitted on the $594$ dev questions, the released $789$ minus the $195$ the benchmark designates for evaluation,
with SubEM as the objective. The benchmark's split holds out questions rather than conversations. The fit is one binary choice per question type against a three-point floor, and the headline
result does not use it at all: the single unfitted configuration already leads the
strongest published agent at all four lengths (Table~\ref{tab:memlens}).

Selection raises the four MemLens scores to $54.87$, $46.67$, $44.10$ and $40.51$, gains of
$+3.59$, $+2.56$, $+2.05$ and $+1.54$ over that single configuration, of which only the
$32$K gain is significant under a paired test ($p=0.0156$). It reaches them with the
benchmark's own question-type annotation, deviating only on knowledge update.

\paragraph{The threshold.}
Without a threshold, choosing per subset would chase noise in small subsets. On MemLens the floor barely matters: dropping it adds
answer refusal to the deviating set and changes one length of four, $64$K by $0.51$.

\section{Separating Delivery from Retrieval: Supporting Analyses}

\subsection{One Question End to End}\label{app:endtoend}

Figure~\ref{fig:teaser} shows one question under two conditions. Here it is under five, ordered by
how good retrieval is. The item that answers it is a photograph of two pieces of salmon on a
board; it arrives at rank $1$ of the ten delivered messages, stamped
\texttt{[2024/05/11 (Sat) 19:33]}. Figure~\ref{fig:payload} shows the bytes.

\begin{table}[ht]
\centering\small
\setlength{\tabcolsep}{4pt}
\caption{One MemLens question through five conditions (32K, benchmark judge), with
retrieval improving down the first four rows.}
\label{tab:endtoend}
\begin{tabular}{lll}
\toprule
Condition & What reaches the model & Answer \\
\midrule
\texttt{cap10} & ten retrieved over the text channel, with pixels & ``To the right\dots'' \\
\texttt{img10} & ten retrieved over both channels, with pixels & ``To the right\dots'' \\
\texttt{oracle10} & the gold messages padded to ten, with pixels & ``To the right\dots'' \\
\texttt{goldonly} & the gold messages alone, no distractors & ``To the right\dots'' \\
\midrule
\texttt{oracle10\_noimg} & the gold messages padded to ten, pixels withheld & ``On top\dots'' \\
\bottomrule
\end{tabular}
\end{table}

Table~\ref{tab:endtoend} follows the salmon question of Section~\ref{sec:intro},
whose gold string is ``To the right''. Four conditions that differ in which messages are delivered, from
ordinary retrieval to the gold messages alone, all answer ``To the
right of the main fillet'' and are scored $1$. The fifth keeps the gold messages padded to ten, withholds the pixels and is scored $0$. Against the third row nothing about the retrieved set changed, and nothing about the question did.

\subsection{Prompts}\label{app:prompts}

Every string in typewriter type below is copied from the code that produced the runs;
text in italics describes an image or content the box leaves out.

\paragraph{The request for one MemLens question.}
We use the benchmark's evaluator prompt verbatim and place the delivered items between its head
and its tail, keeping its refusal instruction and the question date. The box below is the
request for the salmon question in the order the model receives it, with one of the ten delivered
messages written out. The right-hand column names where each block comes from.

\begin{promptbox}{The request for the salmon question, DeliverMem on MemLens}
\scriptsize
\begin{tabular}{@{}p{0.78\linewidth}@{\hspace{8pt}}l@{}}
\ttfamily Provide answers based on the given conversation history. If the question cannot be
answered based on the given conversation, respond with "Insufficient information".
Conversation: & benchmark \\[4pt]
\ttfamily Each recalled item carries its own tag (M1, M2, ...). The tag gives the relevance
rank: M1 is the most relevant, M2 the next, and so on. Identify a memory by its tag, not by
counting items. & (D2) \\[4pt]
\ttfamily [2024/05/11 (Sat) 19:33] Trying to get this recipe write-up done before I crash
tonight. I took a photo while I was testing my citrus-cured salmon and \dots & (D3), message \\[4pt]
\textit{the stored photograph, with M1 stamped in its top-left corner (Figure~\ref{fig:payload})}
& (D1), (D2) \\[4pt]
\textit{nine more retrieved messages, each delivered in the same way} & \\[4pt]
\ttfamily Directly output the answer with no extra output. Question Date: <date> Question:
Where does the smaller chunk of salmon sit compared to the main fillet? Answer with a short
phrase only, no explanation. & benchmark \\
\end{tabular}
\end{promptbox}

\paragraph{(D2) and its delivery-matched control, on MemLens.}
The two conditions deliver identical items, item text, order and budget.
They differ in whether the mark on each item is unique and therefore usable as an address.

\begin{promptbox}{(D2) present, MemLens}
\scriptsize
\begin{verbatim}
  Each recalled item carries its own tag (M1, M2, ...). The tag gives the
  relevance rank: M1 is the most relevant, M2 the next, and so on.
  Identify a memory by its tag, not by counting items.
\end{verbatim}
\end{promptbox}

\begin{promptbox}{(D2) removed, everything else held}
\scriptsize
\begin{verbatim}
  Each recalled item carries a marker (MM). The marker is the same on
  every recalled item and is not part of the content shown. The items
  are listed with the most relevant first.
\end{verbatim}
\end{promptbox}

The control is written under two constraints. It never tells the model that the marker
is useless, since the condition would then measure active discouragement instead of a missing
address. Its last sentence keeps the rank information without making it
addressable, which leaves addressability as the only difference.

\paragraph{(D2) and its control on DMV-Bench.}
The control keeps the first, fourth and fifth sentences verbatim and replaces only the two
that describe addressing, with sentences of similar length.

\begin{promptbox}{(D2) and (D3) present, DMV-Bench: the memory block at one step}
\scriptsize
\begin{verbatim}
  n item(s) recalled from earlier sessions, most relevant first.
  Each recalled image carries its own tag (M1, M2, ...) printed in its
  top-left corner. Identify a memory by the tag you can read in the
  image, not by counting images. Each item also states when you saw
  it: [seen #k of N] means it was the k-th product you ever viewed, so
  a smaller k is longer ago. The current page is shown separately and
  is untagged.
    (1) [seen #k of N] <item text> [tagged M1 in its top-left corner]
    ...
    (5) [seen #k of N] <item text> [tagged M5 in its top-left corner]
\end{verbatim}
\end{promptbox}

\begin{promptbox}{(D2) removed, everything else held}
\scriptsize
\begin{verbatim}
  n item(s) recalled from earlier sessions, most relevant first.
  Each recalled image carries a marker (MM) printed in its top-left
  corner. The marker is the same on every recalled image and is not
  part of the product shown. Each item also states when you saw it:
  [seen #k of N] means it was the k-th product you ever viewed, so a
  smaller k is longer ago. The current page is shown separately and
  is untagged.
\end{verbatim}
\end{promptbox}

Both controls keep rank order available and remove only the address, which makes the two benchmarks' (D2) rows in Table~\ref{tab:controls} comparable.

\paragraph{The tag-in-text variant.} The contrast of Appendix~\ref{app:channel} moves the
tag out of the pixels. Its two altered sentences, in place of the second and third above, are

\begin{promptbox}{(D2) in the text, DMV-Bench}
\scriptsize
\begin{verbatim}
  Each recalled image is introduced by its own tag (M1, M2, ...)
  written in the line just above it. Identify a memory by that tag,
  not by counting images.
\end{verbatim}
\end{promptbox}

\noindent and each item is emitted as \texttt{(i) [seen \#k of N] [M1] <text>} with the image
byte-identical to the stored one. The first, fourth and fifth sentences are those of the
deployed variant, unchanged.

Because the item text and the values of $n$, $k$ and $N$ on DMV-Bench depend on the browsing history,
the box leaves them as placeholders. Figure~\ref{fig:failcase} shows the five images of one such block without their stamps; DeliverMem stamps \texttt{M1} to \texttt{M5} into their top-left corners, as Figure~\ref{fig:payload} shows for MemLens.

\subsection{Content or Presence: Counterfactual Images}\label{app:placebo}

The message set is identical across \texttt{oracle10} and \texttt{oracle10\_noimg}, but
the conditions still differ in how many image tokens the model receives and whether its visual
pathway is engaged at all. Separating those two things from content takes a
ladder of delivery-matched counterfactual images, run as a single batch so that every pair is comparable
question by question. Each counterfactual condition delivers an identical set of image slots and differs only in what the pixels
carry.
\texttt{oracle10\_scramble} permutes the quarter-scale tiles of each delivered image.
Whole objects stay legible, and only their positions are lost.
\texttt{oracle10\_decoy} substitutes a photograph drawn from a different question, sampled
without replacement so that the delivered images remain pairwise distinct.
\texttt{oracle10\_phasescram} randomises the Fourier phase of each delivered image and
then rank-remaps the result. The per-channel pixel histogram is preserved exactly and the
amplitude spectrum to within that monotone map, while no object survives.
The correction for image presence is taken within this batch (Appendix~\ref{app:batches}).

\begin{table}[t]
\centering
\caption{The counterfactual-image ladder on MemLens (32K, benchmark judge, $n=173$ non-refusal
questions, five conditions run and paired within one batch). $\Delta$ is
\texttt{oracle10} minus the condition, so a larger $\Delta$ means a larger loss. The shaded row, the phase-randomised image, keeps the most of the original; Section~\ref{sec:decomp} reports it.}
\label{tab:placebo}
\small
\begin{tabular}{llrrl}
\toprule
Condition & What the delivered pixels carry & $\Delta$ & W/L & $p$ \\
\midrule
\texttt{oracle10\_scramble} & objects legible, positions permuted & $+6.94$ & 17/5 & $0.017$ \\
\rowcolor{black!7}
\texttt{oracle10\_phasescram} & same statistics, no objects & $+11.56$ & 28/8 & $0.0012$ \\
\texttt{oracle10\_decoy} & another question's photograph & $+15.03$ & 33/7 & $4.2\times10^{-5}$ \\
\texttt{oracle10\_noimg} & nothing & $+15.03$ & 34/8 & $6.9\times10^{-5}$ \\
\midrule
\multicolumn{5}{l}{Between counterfactual images} \\
\texttt{decoy} $-$ \texttt{noimg} & a real photograph versus none & $+0.00$ & 6/6 & $1.00$ \\
\texttt{phasescram} $-$ \texttt{decoy} & the correct image's own statistics & $+3.47$ & 9/3 & $0.15$ \\
\bottomrule
\end{tabular}
\end{table}

Table~\ref{tab:placebo} gives the ladder.
A real but wrong photograph scores exactly what no photograph scores, $+0.00$ with $6$
questions moving each way and $p=1.00$. Image tokens and an engaged visual pathway, by
themselves, contribute nothing on this benchmark. Supplying the correct image's own colour
histogram and amplitude spectrum, with every object destroyed, recovers $+3.47$ and falls
short of significance. The effect depends on content. We report
$\texttt{oracle10}-\texttt{oracle10\_phasescram}=+11.56$ as the content-specific value of
original-modality delivery. Of the three conditions that remove the image content, no image, another question's photograph and the phase-randomised image, the last matches the most properties of the original and gives the smallest estimate.

The sixteen-tile permutation is the control for spatial arrangement. MemLens photographs contain a few discrete objects, and a quarter-scale tile is often large enough for a whole one, and a scrambled image keeps many of them intact and relocated. The scrambled images beat the decoys by $+8.09$: both place an image of identical size in identical slots, and the margin is content that survives the shuffle.
We use \texttt{scramble} as a control for spatial arrangement, where
correct positions are responsible for $+6.94$ at $p=0.017$, and the phase-randomised
condition as the control for semantics.

\subsection{Where the Effect Concentrates}\label{app:concentration}

The aggregate delivery effect in this batch can describe two very different things: a
moderate effect on most questions, or a large effect on some and none on the rest.
Splitting that five-condition batch by the benchmark's own question types distinguishes
the two.

\begin{table}[t]
\centering
\caption{Delivering the correct pixels rather than none, by question type (MemLens 32K,
benchmark judge, the batch of Table~\ref{tab:placebo}, paired per question).}
\label{tab:bytype}
\small
\begin{tabular}{lrrrl}
\toprule
Question type & $n$ & $\Delta$ & W/L & $p$ \\
\midrule
Information extraction & 61 & $+40.98$ & 28/3 & $4.6\times10^{-6}$ \\
Multi-session reasoning & 35 & $+2.86$ & 2/1 & $1.00$ \\
Temporal reasoning & 48 & $+0.00$ & 1/1 & $1.00$ \\
Knowledge update & 29 & $+0.00$ & 3/3 & $1.00$ \\
\midrule
Non-refusal (pooled) & 173 & $+15.03$ & 34/8 & $6.9\times10^{-5}$ \\
\bottomrule
\end{tabular}
\end{table}

The effect lives in one of the four types. On questions that ask the model to read a
particular fact off a particular image, such as ``what shape is the Mount Fuji detail
in the artwork my Tokyo friend was studying'', the effect is $+40.98$, and $28$ questions
improve while $3$ worsen. On the remaining three types it is $+2.86$, $+0.00$ and $+0.00$.
Those three ask the model to combine what it saw across several sessions, to notice that
an earlier fact has since been superseded, or to compare two spans of time, and on this backbone delivering the photograph barely moves them. The surviving effect has
$p=4.6\times10^{-6}$, three orders of magnitude below the Bonferroni threshold of $0.0125$
for four comparisons, and two of the three that do not survive are exact zeros.

The aggregate is a large effect on the $35\%$ of questions a picture can answer, diluted by
the $65\%$ that ask something a picture cannot. The two benchmarks in this paper lie at
opposite ends of that mix, since every DMV-Bench probe asks the model to recover one
particular observed item. Section~\ref{sec:boundary} uses this contrast to organise the
component-by-benchmark results.

\subsection{Why Perfect Retrieval Changes So Little}\label{app:coverage}

Section~\ref{sec:decomp} reports that our retriever recovers only a fifth of the gold messages
a question needs, yet perfect retrieval gains just $2.31$ points. Table~\ref{tab:coverage} splits those $173$ non-refusal questions two ways to show why.

\begin{table}[ht]
\centering
\caption{The two quantities of Table~\ref{tab:decomp} by question type and by number of
gold messages (MemLens 32K, its $173$ non-refusal questions, paired within one run). $\Delta_{R}$ is the gold messages minus our retriever, both with pixels, and $\Delta_{D}$ is the gold messages with pixels minus without. The last row reproduces
Table~\ref{tab:decomp}.}
\label{tab:coverage}
\small
\begin{tabular}{lrrrrr}
\toprule
Subset & $n$ & $\Delta_{R}$ & W/L & $\Delta_{D}$ & W/L \\
\midrule
Information extraction & $61$ & $-1.64$ & 5/6 & $+40.98$ & 27/2 \\
Multi-session reasoning & $35$ & $+5.71$ & 2/0 & $-2.86$ & 1/2 \\
Temporal reasoning & $48$ & $+0.00$ & 4/4 & $+2.08$ & 1/0 \\
Knowledge update & $29$ & $+10.34$ & 6/3 & $-3.45$ & 2/3 \\
\midrule
Gold messages $5$--$9$ & $133$ & $+4.51$ & 15/9 & $+18.80$ & 31/6 \\
Gold messages $\ge 10$ & $38$ & $-2.63$ & 2/3 & $-2.63$ & 0/1 \\
\midrule
\rowcolor{black!7}
Non-refusal (pooled) & $173$ & $+2.31$ & 17/13 & $+13.87$ & 31/7 \\
\bottomrule
\end{tabular}
\end{table}

The median question has eight gold messages. The type where delivery matters asks the
model to read one fact off one image, and it needs only the message containing that image.
Our retriever already finds that message, and the seven the oracle adds go unused. On those
$61$ questions delivery is $+40.98$ and perfect retrieval $-1.64$. The oracle's
gains fall on the types that combine evidence across messages. Split by gold count, the pattern repeats: neither axis moves the $38$
questions that need ten or more. Two questions have four gold messages or fewer and are
omitted from the lower block. Coverage and accuracy diverge because a question uses little
of the coverage it is scored on.

No row of Table~\ref{tab:coverage} shows a significant difference between the two retrieval conditions,
every $\Delta_{R}$ having $p\ge0.31$, while $\Delta_{D}$ reaches $p<10^{-4}$ on information
extraction and on questions with five to nine gold messages. On all $173$ non-refusal questions, removing every distractor (the gold messages alone against the padded gold set) and adding our own image channel (both channels against the text channel) each gain $+4.05$, with win/loss counts of 11/4 and 14/7. The two steps cancel in the total headroom, the gold messages alone against retrieval over both channels, which is again $+2.31$, at 18/14.

\subsection{Failure Analysis on DMV-Bench}\label{app:audit}

The MemLens decomposition isolates delivery by ablating it. On DMV-Bench we audit failures
instead.

We instrumented every failed probe on Qwen2.5-VL-7B at $J{=}5$ and asked four questions in order.
Was the target retrieved at all? Retrieval recall is $98.4\%$ for the baseline and $100\%$ for
ours: retrieval leaves little to recover.
Given that it was retrieved, was it in the injected set the model saw? For $76\%$ of the
baseline's failures and $98\%$ of ours, yes: the correct item was in context and a neighbour was chosen. Did the model know what it was looking for? In $93.2\%$ and $94.0\%$ of those
cases the agent's own reasoning trace names the target's object and colour correctly
before selecting the wrong item. Finally, is this a text-to-action mapping error? No: where the trace names an index
outright, that index and the item opened agree in $95$ to $98\%$ of those cases.

On this backbone the failures are therefore selection errors rather than retrieval
errors: $76\%$ of the baseline's and $98\%$ of DeliverMem's, whose failures are fewer. On
Gemini~2.5~Flash, $92\%$ of the baseline's failures are retrieval misses, and the margin comes from (R1) (Appendix~\ref{app:gemini2x2}).

An index misalignment between the model's enumeration of the injected set and ours would produce this signature, and stratification rules it out. For targets in third
position an off-by-one enumeration would give one constant wrong index, $2$ or $4$; the model's reports spread over $2$, $1$ and $3$ instead. The remaining errors are attributions to the wrong item in context, and (D2) gives each item a readable name.

The wrong picks concentrate on one position. When either system fails
this way it most often takes the first delivered item, while the target is first in only $11$ to
$17\%$ of those cases. Because the delivered text is enumerated, position already signals
rank, and a model with nothing else to point with falls back on it.
Appendix~\ref{app:failcase} gives worked cases.

Auditing seven published agents, MemLens's authors find two that retrieve the evidence at
recall $0.82$ to $0.89$ and
still answer $87$ to $95\%$ of their questions wrong after retrieval succeeded,
and attribute this to a backbone unable to reason over the surfaced content
\citep{memlens}. With the backbone and the surfaced messages unchanged, changing only the
form in which the messages arrive improves accuracy by $13.87$ points. The apparent limit on reading comprehension is in part a property of the interface.

\section{Additional Results}

\subsection{Backbone Scale}\label{app:scale}

All three backbones run one protocol, with the retrieval index, $K$ and judge fixed and only the answering model changed, from Qwen3-VL-8B to Qwen3-VL-30B-A3B to Qwen3-VL-32B. The two larger models differ in kind as well as in size. Qwen3-VL-30B-A3B is a mixture of experts with $30$B total and $3$B active parameters: more total capacity than the $8$B dense model and less active capacity. Table~\ref{tab:scale} orders the backbones by total parameters. The gap between the two terms is itself a paired contrast, since $\Delta_{D}-\Delta_{R}=A(R,D)-A(R^{\star},D_{0})$, our retriever with pixels against the gold messages without them. It is $+11.56$ at $8$B ($36$/$16$, $p=0.0078$), $+5.78$ at 30B-A3B ($28$/$18$, $p=0.18$) and $+6.36$ at $32$B ($28$/$17$, $p=0.14$).

On the questions it answers, the $32$B model is more accurate than the $8$B model, $61.74\%$
against $57.66\%$, although its overall score in this batch, with ten messages retrieved over both
channels, is $47.69$ against $51.79$. The difference is abstention: on the $173$ answerable questions the
$32$B model declines to answer $58$ of them and the $8$B model $36$, and a declined
answerable question is scored wrong. Of the $24$ questions the $32$B model declines and
the $8$B model attempts, the $8$B model answers $12$ correctly. The reverse set has $2$
questions, and the $32$B model answers neither. Those $12$ questions exceed the whole gap
between the two models, $8$ questions of the $195$. With both retrieval channels the abstention rate is
$20.8\%$ at $8$B, $12.1\%$ at 30B-A3B and $33.5\%$ at $32$B, and it tracks the delivered
evidence, falling on the $32$B model from $36.4\%$ with the text channel alone to $22.0\%$ with
gold messages alone. Abstention does not produce the ordering of the delivery term: 30B-A3B declines least and still has a larger delivery term than the $8$B model.

\paragraph{Metric dependence of the delivery:retrieval ratio.}
Under SubEM the $32$B delivery:retrieval ratio is $1.71\times$ against the judge's $1.55\times$, with
the retrieval term at $+9.83$ under SubEM and $+11.56$ under the judge, both $p<0.002$. On 30B-A3B
SubEM puts the retrieval term at $+5.78$ ($21$/$11$, $p=0.11$) where the judge puts it at $+9.83$.
Only the paired differences compare across the three rows.

\begin{table}[ht]
\centering
\caption{The two quantities of Section~\ref{sec:decomp} on three Qwen3-VL backbones (MemLens 32K,
$n=173$ non-refusal questions, benchmark judge, paired within run). Delivery is the gold messages with pixels minus without, and retrieval is the gold messages minus our retriever, both with pixels. Backbones are ordered by total parameters.}
\label{tab:scale}
\small
\begin{tabular}{llrrl}
\toprule
Backbone & Quantity & $\Delta$ & W/L & $p$ \\
\midrule
$8$B dense  & Delivery  & $+13.87$ & 31/7  & $1.2\times10^{-4}$ \\
      & Retrieval & $+2.31$  & 17/13 & $0.58$ \\
\midrule
30B-A3B & Delivery  & $+15.61$ & 36/9  & $6.6\times10^{-5}$ \\
 (MoE)     & Retrieval & $+9.83$  & 21/4  & $9.1\times10^{-4}$ \\
\midrule
$32$B dense & Delivery  & $+17.92$ & 32/1  & $7.9\times10^{-9}$ \\
      & Retrieval & $+11.56$ & 26/6  & $5.4\times10^{-4}$ \\
\midrule
\multicolumn{2}{l}{Ratio delivery\,:\,retrieval} & \multicolumn{3}{l}{$8$B\ \ $6.00\times$\qquad 30B-A3B\ \ $1.59\times$\qquad $32$B\ \ $1.55\times$} \\
\bottomrule
\end{tabular}
\end{table}

\subsection{Same-Batch Paired Comparison on DMV-Bench}\label{app:paired}

\begin{table}[ht]
\centering
\caption{Same-batch paired comparison against our own run of DualMem at its strongest
setting. Estimates are seed-averaged, and $p$ is an exact cluster permutation whose floor
is $0.002$ at $k{=}10$ chains and $0.031$ at $k{=}6$.}
\label{tab:dmv_paired}
\small
\begin{tabular}{llrrrrl}
\toprule
Backbone & Setting & Same-batch & DeliverMem & $\Delta$ & same dir. & $p$ \\
\midrule
\multirow{5}{*}{Qwen2.5-VL-7B}
 & $J{=}5$ & 79.88 & 87.51 & $+7.63$ & 8/10 & $0.020$ \\
 & $J{=}10$ & 77.80 & 89.55 & $+11.75$ & 10/10 & $0.002$ \\
 & $J{=}15$ & 76.59 & 88.94 & $+12.34$ & 10/10 & $0.002$ \\
 & $J{=}50$ (exh.) & 72.34 & 85.20 & $+12.86$ & 10/10 & $0.002$ \\
 & $J{=}50$ (MC) & 72.17 & 84.96 & $+12.79$ & 10/10 & $0.002$ \\
\midrule
\multirow{4}{*}{Gemini 2.5 Flash}
 & $J{=}5$ & 80.68 & 93.26 & $+12.58$ & 6/6 & $0.031$ \\
 & $J{=}10$ & 73.73 & 90.30 & $+16.57$ & 6/6 & $0.031$ \\
 & $J{=}15$ & 70.70 & 89.45 & $+18.75$ & 6/6 & $0.031$ \\
 & $J{=}50$ (MC) & 65.99 & 88.61 & $+22.62$ & 6/6 & $0.031$ \\
\bottomrule
\end{tabular}
\end{table}

At $J{=}50$ the benchmark scores a Monte-Carlo sample of probes rather than every one: ten probes per recall reach per chain, where a probe's reach is the number of sessions between seeing a product and being asked for it. The exhaustive construction scores every probe, about ten times as many (Table~\ref{tab:published}). The Monte-Carlo row gives $+12.79$ and the exhaustive row $+12.86$, and the main table reports the sampled row for protocol parity.

\subsection{The Same $2\times2$ on Gemini 2.5 Flash}\label{app:gemini2x2}

Section~\ref{sec:controls} decomposes the $J{=}50$ margin on Qwen2.5-VL-7B. These four cells on Gemini~2.5~Flash place almost none of it on delivery. Six chains contribute,
scored on the $2{,}940$ probes common to all four.

\begin{table}[ht]
\centering
\caption{The $2\times2$ at $J{=}50$ on Gemini~2.5~Flash. Task success rate (\%) over the
probes common to all four cells. The delivery column adds the tag and the ordinal together.}
\label{tab:gemini2x2}
\small
\begin{tabular}{lrr}
\toprule
Retrieval & delivery off & delivery on \\
\midrule
DualMem & $65.99$ & $66.29$ \\
(R1) region multi-vector & $87.99$ & $\mathbf{88.61}$ \\
\bottomrule
\end{tabular}
\end{table}

In Table~\ref{tab:gemini2x2}, (R1) alone moves accuracy by $+22.01$ and delivery alone by
$+0.31$; together they give the margin Table~\ref{tab:dmv_paired} reports. The main effects are $+22.16$ for (R1), with all six
chains agreeing in sign at the permutation floor $p=0.031$, and $+0.46$ for delivery at
$p=0.156$. The interaction is $+0.31$ ($p=0.656$).

The failure audit of Appendix~\ref{app:audit} anticipates the split: on Gemini nearly all of
the baseline's failures are retrieval misses, whereas on Qwen they are selection errors among
items already in context. Delivery acts on what the model has, and where the item never arrives (R1) takes the whole margin. Which stage pays depends on what the backbone fails at rather than on the benchmark alone.

Delivery's absolute gain is bounded by how much is left to win, so the null has to be
told apart from a ceiling. Taking each delivery gain as a share of the points still
available above its own retrieval condition, delivery recovers $30.3\%$ of that headroom on
Qwen2.5-VL-7B and $5.1\%$ here. The gap survives the normalisation, so the ceiling does
not account for it.

A lighter memory load does not bring the term back. Those four cells at $J{=}10$, six chains
over the $3{,}079$ probes common to them, give (R1) a main effect of $+16.30$ at the
permutation floor and delivery $+0.27$ at $p=0.313$; delivery recovers $4.8\%$ of the
headroom above its own retrieval condition, against $5.1\%$ at $J{=}50$, so the null belongs to the backbone rather than to the working point.

\subsection{Margin Growth with Memory Load}\label{app:scaling}

\begin{figure}[t]
\centering
\includegraphics[width=\linewidth]{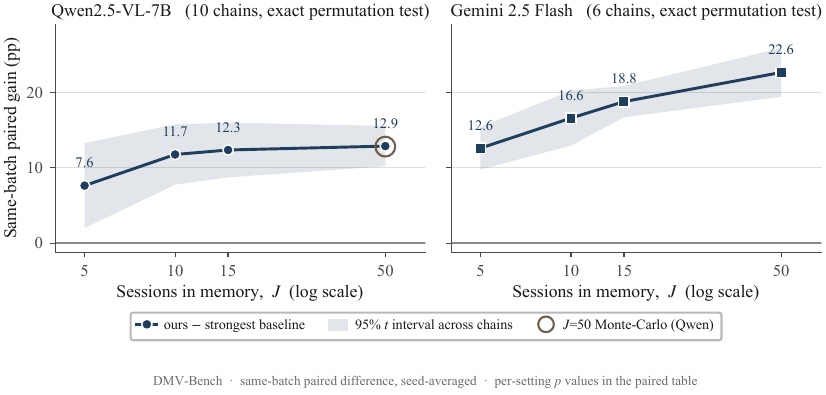}
\caption{Paired difference against the same-batch DualMem run as memory load grows, for
Qwen2.5-VL-7B ($k{=}10$ chains, left) and Gemini 2.5 Flash ($k{=}6$, right). Bands are
$95\%$ $t$ intervals across chains. At $J{=}50$ Qwen's point is the exhaustive construction
with the Monte-Carlo value ringed, and Gemini was run under Monte-Carlo only.}
\label{fig:scaling}
\end{figure}

The margin widens with memory load because the opponent degrades faster than we do: over that range its score falls by more than three times as much as ours on both backbones
(Table~\ref{tab:dmv_paired}). Figure~\ref{fig:scaling} plots the paired difference against memory
load for both.

\subsection{Delivery Budget}\label{app:knobs}

We deliver $K{=}10$ retrieved messages. Sweeping it (benchmark judge, $32$K) gives
$40.00$ at $K{=}3$, $46.67$ at $K{=}5$, $51.28$ at $K{=}10$ and $51.28$ at $K{=}20$. The
curve climbs to ten and stops there. Going from three to ten gains $+11.28$, and going from
ten to twenty changes the judge score by exactly zero.
$K{=}10$ is fixed everywhere else on MemLens, and DMV-Bench delivers the five its own
protocol specifies. Since doubling the delivered set changes nothing at 32K, a larger budget would buy no accuracy there.

\subsection{Input Cost}\label{app:cost}

We compare memory-agent accuracy and input cost against direct inference with our backbone, in which the entire conversation goes to a long-context
model. Token counts are computed offline with the backbone's own tokenizer for
text. For images we skip decoding and apply the model's own resize rule to the
stored $(h,w)$.\footnote{Qwen3-VL encodes $16\times16$ patches and merges them in $2\times2$ groups; one visual token covers $32\times32$ pixels.} Direct inference is budgeted at our own reduced image resolution, the longer side capped at $768$ pixels. A long-context system would send larger images, and the ratios
below understate the gap. The ratio counts only what reaches the answering model, leaving out the offline work behind it: (R1) encodes every stored image as a whole vector plus a
$2\times2$ overlapping grid, five vectors per image, and scores them by MaxSim. That cost
is paid once per stored item rather than once per question, and we measure it here.
Five vectors take five times the storage of one but not five times the work: the five crops share one decode and one batched
forward pass, so encoding an image costs $1.9\times$ a whole-image encoding on DMV-Bench
and about $3.5\times$ on MemLens, whose smaller photographs make decoding a smaller
share of each call. A question's conversation holds a median of $13$ images at 32K and
$96$ at 256K, an index of $0.20$ to $1.47$\,MB built in under four seconds on one
$32$\,GB GPU. Images recur across questions, so the index over an entire length is
$4{,}090$ to $4{,}686$ distinct images, $63$ to $72$\,MB, under three minutes. A
DMV-Bench chain stores fewer, $21$ images at $J{=}5$ and $80$ to $157$ at $J{=}50$
depending on how much of the storefront the backbone browses. Per query, scoring five vectors per image over the 128K index takes about $2$\,ms, against $0.15$\,ms for one vector per image. We measure on
$200$ sampled images per corpus.

\begin{table}[t]
\centering
\caption{Input cost and accuracy against direct inference with the same backbone.
``Input ratio'' is median full-context tokens over median delivered tokens. The benchmark runs
direct LVLMs at three lengths only, since many of them do not reach $256$K, so the last
column has no direct-inference reference for any model \citep{memlens}.}
\label{tab:cost}
\small
\begin{tabular}{lrrrr}
\toprule
& 32K & 64K & 128K & 256K \\
\midrule
Our delivered input (tokens) & 2{,}405 & 2{,}421 & 2{,}460 & 2{,}450 \\
Full context (tokens)        & 24{,}001 & 46{,}476 & 89{,}940 & 178{,}558 \\
Input ratio                  & $10.0\times$ & $19.2\times$ & $36.6\times$ & $72.9\times$ \\
\midrule
DeliverMem                          & \textbf{51.28} & 44.10 & \textbf{42.05} & 38.97 \\
Direct inference, same backbone     & 50.77 & \textbf{47.69} & 34.36 & n/a \\
Accuracy difference                 & $+0.51$ & $-3.59$ & $+7.69$ & n/a \\
\bottomrule
\end{tabular}
\end{table}

\paragraph{Stability of delivered input size.}
The delivered size reported in Section~\ref{sec:results} stays near-constant while its composition shifts: the text component falls
$24\%$ and the visual component rises $127\%$, as longer haystacks bring more
image-bearing messages into the top ten, from $3$ to $7$ images.

\subsection{Temporal Position across Context Lengths}\label{app:d3len}

The effect stays confined to temporal questions at every
context length: on the $195$-question subset the contrast gains $+10.42$, $+8.33$, $+8.33$
and $+10.42$ on temporal questions at the four lengths while the net outside them is
$+1.36$, $-2.72$, $-0.68$ and $-0.68$.

In the run that includes no date, accurate dates and permuted dates, accurate dates improve on no date by $8.76$ points on the $194$ temporal questions ($28$/$11$, $p=0.0095$) and on permuted dates by $14.43$ ($32$/$4$, $p=1.9\times10^{-6}$), and permuted dates fall $5.67$ below no date ($14$/$3$, $p=0.013$). 

\subsection{Per-Type Effects of the Assembled Configuration}\label{app:subgroup}

Across the full MemLens question set, the tag, the date and (R1) together cost $-8.62$ points on
knowledge update while gaining $+7.22$ on temporal reasoning. The dev-split rule therefore defaults to the baseline on knowledge update. The cost belongs to no single component, but
the gain does: on temporal reasoning the date contributes $+9.28$ and the tag
$-2.06$. These figures use all $789$ released questions. On that set no single delivery
variant has a cost specific to knowledge update.

Within the final batch of Table~\ref{tab:pool} in Appendix~\ref{app:pool}, adding the tag
and the date moves accuracy by $+1.03$ points overall, a net of two questions: seven gained on temporal
reasoning ($+14.58$), four lost on knowledge update ($-13.79$), two lost on information extraction ($-3.28$) and one gained on multi-session reasoning ($+2.86$).

\subsection{What (R1) Depends On}\label{app:region}

\begin{figure}[ht]
\centering
\includegraphics[width=\linewidth]{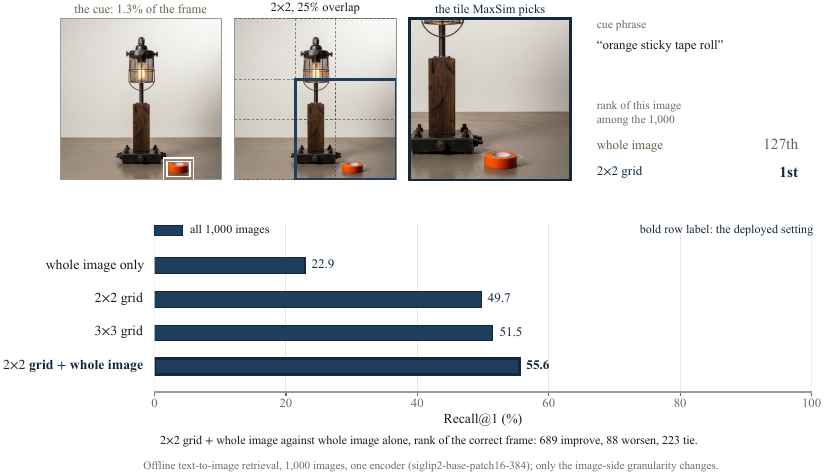}
\caption{(R1) on DMV-Bench's cued frames. Top: one frame with its inserted cue, the overlapping tiles and the tile MaxSim scores highest. Bottom: offline Recall@1 at four
granularities with one encoder, the deployed setting in the bottom row.}
\label{fig:region}
\end{figure}

(R1) stores each image as a whole plus an overlapping grid and scores an item by its
best-matching region. Whether that helps is a property of the corpus. DMV-Bench inserts its cue with a templated instruction asking for an object
``subtle and modest in size, clearly visible but not dominating the scene'', leaving the
scene otherwise unchanged \citep{dmvbench}. Differencing each cued frame against its own
uncued original isolates a single object in $847$ of the $1{,}000$ frames, and there the
object occupies a median $1.7\%$ of the image. Whole-image pooling averages an object
that small into the vector of the furniture it sits on.

Figure~\ref{fig:region} scores all $1{,}000$ frames against their own cue phrases with
one encoder, changing only the image-side granularity. Recall@1 rises from $22.9$ to $55.6$, and the rank of the correct frame improves for $689$ frames and worsens for $88$. The frame in the top row ranks $127$th of $1{,}000$ for its own cue phrase under the
whole-image encoding and first under the $2\times2$ encoding. On the $847$ frames with an
isolated cue, the region MaxSim selects contains that object $99.2\%$ of the time
against $60.3\%$ for a tile drawn at random, $99.5\%$ when the frame is retrieved at rank one
and $98.8\%$ when it is not. The failure is in the ranking.

\subsection{Binding Control by Question Type}\label{app:binding}

\begin{table}[ht]
\centering\small
\caption{The binding control by MemLens question type (32K, benchmark judge).
The control permutes only which image follows which message (Appendix~\ref{app:arms}), and $\Delta$ is paired within one run.}
\label{tab:binding_type}
\begin{tabular}{lrrrrl}
\toprule
Question type & $n$ & Full & Binding shuffled & $\Delta$ & W/L \\
\midrule
Information extraction  & 61 & 59.02 & 50.82 & $+8.20$ & 6/1 \\
Multi-session reasoning & 35 & 22.86 & 20.00 & $+2.86$ & 2/1 \\
Temporal reasoning      & 48 & 50.00 & 43.75 & $+6.25$ & 3/0 \\
Knowledge update        & 29 & 37.93 & 31.03 & $+6.90$ & 3/1 \\
Answer refusal          & 22 & 100.00 & 100.00 & $+0.00$ & 0/0 \\
\midrule
Non-refusal (pooled)    & 173 & 45.66 & 39.31 & $+6.36$ & 14/3 \\
\bottomrule
\end{tabular}
\end{table}

The effect is significant when pooled over the answerable types, and every one of them moves in the same direction (Table~\ref{tab:binding_type}).

\subsection{Equal-Budget Control: Full Table}\label{app:iso}

Table~\ref{tab:iso} delivers ten messages selected without the query through the identical pipeline, which separates query conditioning from the delivery machinery: both control conditions receive an identical token budget and rendering.

\begin{table}[ht]
\centering\small
\caption{Selection without the query (MemLens judge score, canonical $195$ questions).
All rows deliver ten messages through one pipeline and differ only in which ten are
selected.}
\label{tab:iso}
\begin{tabular}{lrr}
\toprule
Selection & 32K & 128K \\
\midrule
Query-conditioned (ours) & 51.28 & 42.05 \\
Recency, no query & 22.05 & 14.36 \\
Random, no query & 17.44 & 14.87 \\
\bottomrule
\end{tabular}
\end{table}

At an identical budget and rendering, removing query conditioning costs $27.69$ to
$29.23$ points against recency selection and $27.18$ to $33.85$ against random selection.

\section{Statistical Detail and Worked Cases}

\subsection{Runs and Batches}\label{app:batches}

Most paired contrasts in this paper compare conditions scored together, on one set of questions or probes with one answering model, prompt template and judge; the rest pair the same questions or probes across two runs and are listed here. Across runs only paired differences compare, and a condition scored in two runs agrees with itself to about one question. Table~\ref{tab:decomp} needs one condition from a second run: retrieval without pixels, scored alongside a re-run of retrieval with pixels that matched the first run on all $173$ non-refusal questions. Every other cell of the table, and both quantities of Figure~\ref{fig:teaser}, come from the first run. The withheld-pixel contrast is $+13.87$ there and $+15.03$ in the counterfactual-image
batch of Table~\ref{tab:placebo}; $\Delta_{D}$ on multi-session reasoning is $-2.86$ ($1$/$2$) in Table~\ref{tab:coverage} and $+2.86$ ($2$/$1$) in Table~\ref{tab:bytype}; and the
permuted-date contrast of Section~\ref{sec:controls} improves $31$ temporal questions against
$4$ in one run and $32$ against $4$ in another. On DMV-Bench the two columns of
Table~\ref{tab:channel2x2} come from different batches, and its contrasts between columns pair the same $1{,}888$ probes across them; the deployed variant, scored in both, differs by $1.09$ points ($p=0.21$), which is the gap between the $+10.97$ there and the $+9.88$ of Table~\ref{tab:controls}. Our MemLens rows in Table~\ref{tab:pool} come from
one run on the default image index and one on the region index of the deployed configuration; Table~\ref{tab:memlens}'s two variant rows retrieve over text alone and come from the first. The baseline of Appendix~\ref{app:routing} comes from a separate run of the default-index condition, one knowledge-update question apart from Table~\ref{tab:pool}'s. The $8$B rows of Table~\ref{tab:scale} are those of Table~\ref{tab:decomp}; its larger backbones, and the absolute scores discussed with it, come from runs of their own, in which the $8$B two-channel condition scores $51.79$ where Table~\ref{tab:pool}'s scores $51.28$. The $26$ questions that restoring the pixels regains in Section~\ref{sec:results} come from the run of Table~\ref{tab:pool}, where text-channel retrieval answers $73$ of the $173$ non-refusal questions; the run of Table~\ref{tab:decomp} answers $72$, a net of $25$. Differences are computed before rounding and can sit $0.01$ from the difference of the rounded values shown. Because the binding control of Table~\ref{tab:binding_type} uses the default index,
its absolute scores do not match Table~\ref{tab:memlens}, while its paired difference, taken
within one run, remains comparable.

\subsection{Seed Variance}\label{app:variance}

Per-seed variance (Table~\ref{tab:variance}) splits into a binomial term (finite probes
per seed) and a residual,
$\mathrm{SD}_{\text{res}}=\sqrt{\mathrm{SD}_{\text{obs}}^2-\mathrm{SD}_{\text{binom}}^2}$.
The binomial term assumes probes are independent within a seed. Because our protocol nests
probes, that assumption can fail, and the residual is an upper bound on chain-to-chain
difficulty rather than an estimate of it.

\begin{table}[ht]
\centering\small
\caption{Per-seed variance on DMV-Bench with Qwen2.5-VL-7B, split into a binomial term
and a residual.}
\label{tab:variance}
\begin{tabular}{lrrrr}
\toprule
Cell & probes/seed & obs.\ SD & binomial SD & residual \\
\midrule
Qwen $J{=}5$  & 43   & 8.01 & 6.09 & 5.20 \\
Qwen $J{=}10$ & 189  & 6.31 & 3.02 & 5.54 \\
Qwen $J{=}15$ & 436  & 5.21 & 2.03 & 4.80 \\
Qwen $J{=}50$ & 4870 & 3.15 & 0.64 & 3.09 \\
\bottomrule
\end{tabular}
\end{table}

At $J{=}5$ a seed contains $43$ probes, and the binomial term alone is $6.09$ points.
The residual adds $3$ to $5.5$ points and shrinks far more slowly than the binomial term as $J$ grows: more probes per seed barely reduce it.

\subsection{Worked Failure Cases}\label{app:failcase}

We give three DMV-Bench failures that satisfy all three conditions of the audit: the
target was retrieved, it was in the injected set, and the agent's own trace names the
cue correctly before it commits to a different item. All three come from the constant-tag control of Table~\ref{tab:controls}, with Qwen2.5-VL-7B at $J{=}10$: every delivered image carries the same symbol and the model has no name to point with. Traces, delivered sets and outcomes are quoted verbatim from the agent traces, and the position of the
target is computed from the delivered list recorded in them.

\begin{figure}[ht]
\centering
\includegraphics[width=\linewidth]{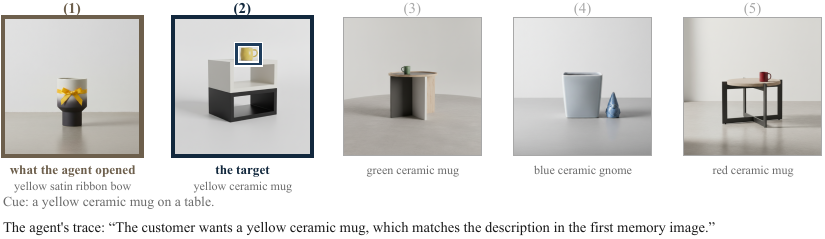}
\caption{The first of the three, with the delivered set shown without its stamps. Five items
reach the model and the target is second. The box marks the cue, located by differencing the frame
against its uncued original.}
\label{fig:failcase}
\end{figure}

\begin{itemize}\itemsep3pt
\item Cue: a yellow ceramic mug on a table. The target is second of five delivered
items; the agent opens the first, a yellow ribbon bow on a vase, although the third and
fifth items also hold ceramic mugs. Its trace reads ``The customer wants a yellow ceramic
mug, which matches the description in the first memory image''. Figure~\ref{fig:failcase}
shows what it was given.
\item Cue: a green eucalyptus sprig in a vase. The target is third of five; the agent opens
the first, which shows the same sprig in blue: ``The recalled product with the green
eucalyptus sprig is the first one in the list''.
\item Cue: a green wool scarf on a chair. The target is second of five; the agent opens
the first, a green cardigan on a chair, and its trace ends ``This matches the first image
in the memory context''.
\end{itemize}

In each case the description is correct and the item opened is not: the trace attributes the cue to the first image, and the agent opens it. This is the form most failures take, the action following the attribution in $95$ to $98\%$ of the audited failures whose trace names an index (Appendix~\ref{app:audit}). With a distinct name on each image, the deployed tag gains $11.85$ points over this control, all ten chains agreeing in sign (Table~\ref{tab:controls}).

On MemLens the corresponding failure has two forms. Withholding the pixels sometimes
produces an abstention: asked for the name of a design studio shown in a
photograph, the model answers correctly with the image and ``Insufficient
information'' without it. Sometimes it produces a confident error, as in the
salmon question of Appendix~\ref{app:endtoend}.

\end{document}